\documentclass[a4paper,fleqn]{cas-dc}

\usepackage[numbers]{natbib}
\usepackage{ragged2e}
\usepackage{algorithm}
\usepackage{algorithmic}
\usepackage{subcaption}

\def\tsc#1{\csdef{#1}{\textsc{\lowercase{#1}}\xspace}}
\tsc{WGM}
\tsc{QE}
\tsc{EP}
\tsc{PMS}
\tsc{BEC}
\tsc{DE}

\begin{document}
\justifying
\let\WriteBookmarks\relax
\def\floatpagepagefraction{1}
\def\textpagefraction{.001}
\shorttitle{\texorpdfstring{SM$^2$C}{} -- Boost the Semi-supervised Segmentation for Medical Image}
\shortauthors{Yifei Wang et~al.}
\let\printorcid\relax


\title [mode = title]{\texorpdfstring{SM$^2$C}{} -- Boost the Semi-supervised Segmentation for Medical Image by using Meta Pseudo Labels and Mixed Images} 

\author[1]{Yifei Wang}[type=editor,
                        role=Researcher,]

\credit{Conceptualization of this study, Methodology, Software}

\address[1]{Department of Biomedical Engineering, Chongqing University, Chongqing 400030, China.}

\author[1,2]{Chuhong Zhu}[style=chinese, role=Corresponding,
                          orcid=0000-0003-4146-5405]

\cormark[1]

\credit{Data curation, Writing - Original draft preparation}

\address[2]{Engineering Research Center for Organ Intelligent Biological Manufacturing of Chongqing, Key Lab for Biomechanics and Tissue Engineering of Chongqing, Third Military Medical University, Chongqing 400038, China.}


\begin{abstract}
Recently, machine learning-based semantic segmentation algorithms have demonstrated their potential to accurately segment regions and contours in medical images, allowing the precise location of anatomical structures and abnormalities. Although medical images are difficult to acquire and annotate, semi-supervised learning methods are efficient in dealing with the scarcity of labeled data. However, overfitting is almost inevitable due to the limited images for training. Furthermore, the intricate shapes of organs and lesions in medical images introduce additional complexity in different cases, preventing networks from acquiring a strong ability to generalize.
To this end, we introduce a novel method called Scaling-up Mix with Multi-Class (SM$^2$C). This method uses three strategies - scaling-up image size, multi-class mixing, and object shape jittering - to improve the ability to learn semantic features within medical images. By diversifying the shape of the segmentation objects and enriching the semantic information within each sample, the SM$^2$C demonstrates its potential, especially in the training of unlabelled data. Extensive experiments demonstrate the effectiveness of the SM$^2$C on three benchmark medical image segmentation datasets. The proposed framework shows significant improvements over state-of-the-art counterparts. 
\end{abstract}


\maketitle

\section{Introduction}
Semantic segmentation is a popular task in computer vision, where the goal is to assign a categorical label to each pixel in an image. In recent years, rapid advances in supervised deep learning methods have led to the development of numerous techniques, including FCN \cite{r2}, U-Net \cite{r1}, and SegNet \cite{r3}. These techniques show remarkable potential for automating image segmentation, a capability with significant implications in several domains, notably medical image analysis \cite{r4, r5}. Given the scarcity of medical resources, such as professional experts and equipment, the application of image processing in medical assistance has received considerable attention. These techniques, particularly image segmentation, not only help medical professionals to locate regions of interest, but also contribute to surgical planning and intra-operative guidance.

However, the effectiveness of current supervised methods for medical image processing is closely related to the availability of accurately annotated training data, which is a critical factor in the performance of a segmentation model on a test dataset\cite{r6,r7}. Unfortunately, the annotation of medical images is a laborious task that requires expertise, resulting in a significant bottleneck. A direct strategy to tackle this challenge is to incorporate unlabeled images with annotated data.

To this end, researchers have turned to semi-supervised learning (SSL) algorithms for medical image segmentation. The aim is to extend medical image datasets with a large number of unlabelled images. Semi-supervised techniques include methods such as pseudo-labelling \cite{r8}, consistency regularisation \cite{r9, r41, r42, r43}, co-training \cite{r6, r10} and so on. Among these, pseudo-labelling methods \cite{r11, r12} have demonstrated efficiency in classification tasks by using unlabelled samples. Several studies \cite{r14, r10, r13} have extended these methods to medical image segmentation to address the challenge of limited labeled data. Pseudo-labeling methods require a teacher model that uses information from labeled data to generate pseudo-labels, which guide a student model to learn from unlabelled data.
However, there is a drawback in this type of framework due to the lack of limited semantic information contained in the labeled data. Therefore, it is important to extend the semantic information in the training of the teacher model by using unlabelled data. In this process, unlike the training procedure of the student model, which is supervised by distilled knowledge from unlabelled data, the teacher model is required to use the semantic information in unlabelled data by unsupervised strategies. Consistency regularisation is effective in unsupervised learning of unlabeled data, by adding in predictions before and after perturbations. Therefore, there is a tendency \cite{r15, r38} to integrate consistency regularisation into pseudo-labeling for medical image segmentation.

\begin{figure}[t]
	\centering
 
	\subcaptionbox{\label{fig:a}} 
	{\includegraphics[width = .21\linewidth]{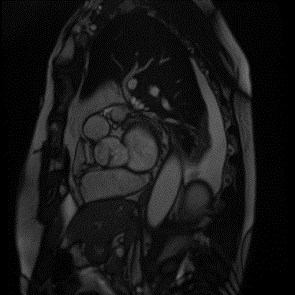}}
        \quad
        \subcaptionbox{\label{fig:b}} 
	{\includegraphics[width = .21\linewidth]{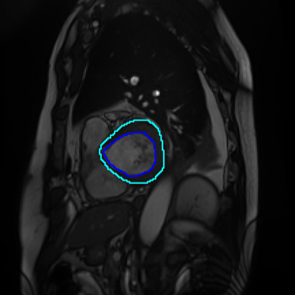}}
        \quad
	\subcaptionbox{\label{fig:c}}
	{\includegraphics[width = .21\linewidth]{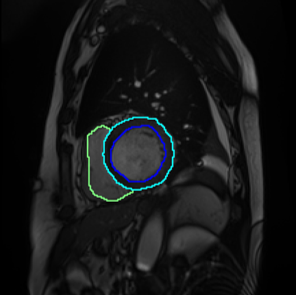}}
        \quad
	\subcaptionbox{\label{fig:d}}
	{\includegraphics[width = .21\linewidth]{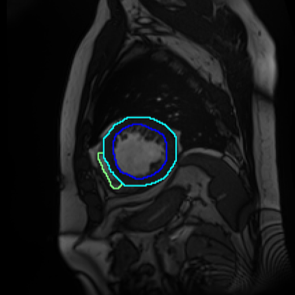}}
 
	\caption{Different sections of a heart. (a) represents the basal section, which lacks regions of the right ventricle(RV), myocardium, and the left ventricle(LV); (b), (c), and (d) approach the apical section, while the RV, myocardium, and LV vary in size and shape.}
	\label{fig:fig8}
\end{figure}

French et al.\cite{r16} emphasized that consistency regularisation requires robust and diverse perturbations in semantic segmentation, which leads networks to possess a strong ability to generalize. Perturbations are usually performed using a set of data augmentation algorithms, and consistency regularization uses highly augmented images to efficiently learn unlabelled data. In terms of the intricate anatomical shapes in greyscale medical images, data augmentation algorithms should be able to enable the segmentation model to identify the shapes of objects, especially contours. Furthermore, medical images sometimes do not contain objects from some specific classes, which can lead to inter-class imbalance (Figure \ref{fig:fig8}). To address these challenges, we introduce Scaling-up Mix with Multi-Class (SM$^2$C), a novel data augmentation algorithm based on image mixing techniques to generate a more diversified training dataset. Targeting the primary tasks of semantic segmentation - region and contour recognition - our algorithm processes the shapes of segmentation objects and merges semantic features from multiple original images into a novel composition. We incorporate the method into a pseudo-labeling framework to improve the teacher model's ability to learn unlabeled data and conduct experiments on three medical image datasets.

 The main contributions of the study are summarized as follows:
\begin{itemize} 
\item A novel data augmentation method called Scaling-up Mix with Multi-Class (SM$^2$C) is introduced. The me-thod improves the diversity of the samples in both image size and object shape dimensions. By integrating it with a popular semantic segmentation framework, the segmentation model pays more attention to the region and contour of objects.
\item The experimental results show that the proposed framework with SM$^2$C is superior to other popular semi-supervised frameworks on three different medical image datasets.
\item Ablation experiments are performed to demonstrate the effects of the three components of the SM$^2$C algorithm on the segmentation model. Furthermore, the different effects of different data augmentation methods are discussed.
\end{itemize}

The rest of this paper is structured as follows. In Section 2, the fundamentals of semi-supervised medical image segmentation are reviewed. Section 3 introduces the proposed shape and boundary-aware data augmentation method. Section 4 discusses the comparative experiments and results. Section 5 gives a general summary of the whole work.

\section{Related work}
\begin{figure} 
	\centering
	\includegraphics[width = .9\linewidth]{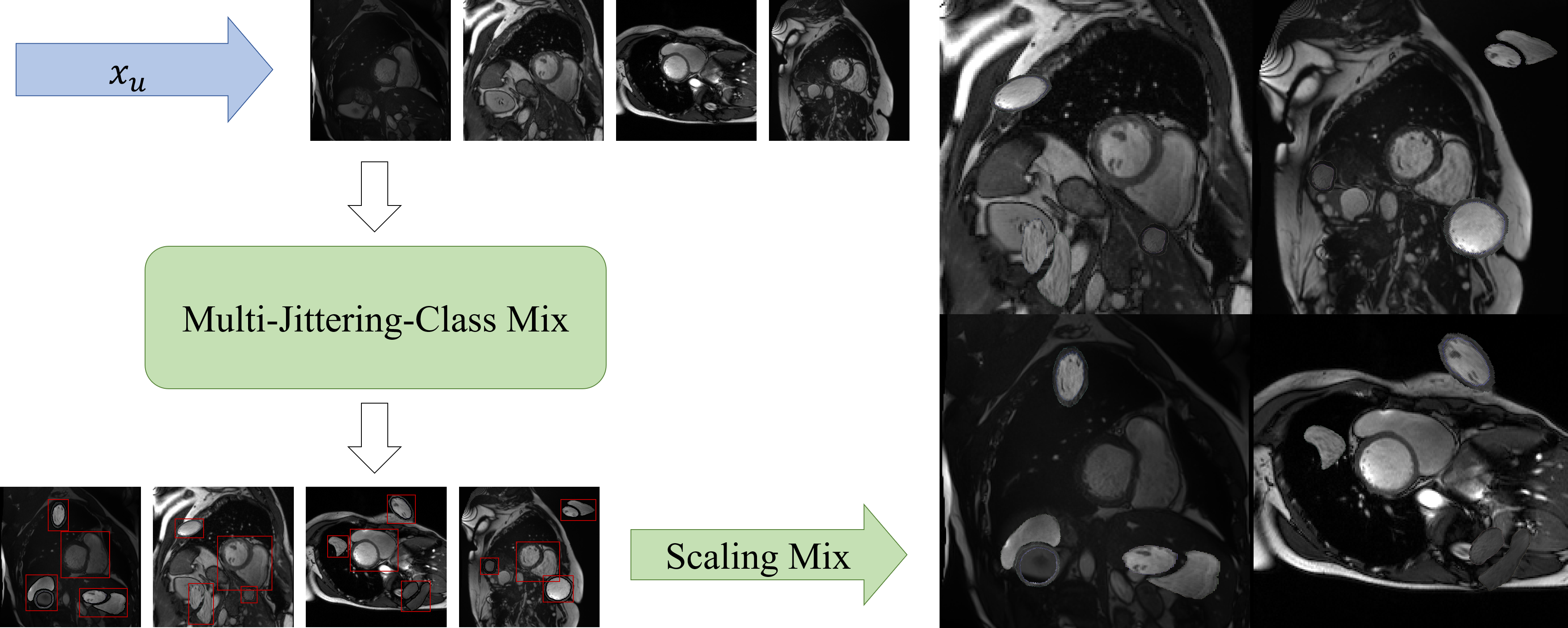}
        \caption{Illustration of Scaling-up Mix with Multi-Class (SM$^2$C). Four unlabeled images (quantity unlimited) extract their respective segmentation objects, augment these objects, randomly mix them with other images, and finally, through a concatenation operation, create a larger-sized input image containing more content.}
	\label{fig:fig1}
\end{figure}

\subsection{Semi-supervised segmentation}


Semi-supervised learning is an effective method to reduce the dependence of deep segmentation models on labeled images. This approach is typically applied to a dataset containing both labeled and unlabelled images. The segmentation models extensively learn the semantic information from the labeled images and perform unsupervised learning on the unlabelled images. Based on the different methods used to learn from unlabelled images, the current mainstream semi-supervised semantic segmentation algorithms can be roughly divided into two groups.

First, there are methods based on consistency regularisation. These methods improve the consistency of the model in predicting the same image under different perturbations to learn from unlabelled images. The perturbation methods are critical to whether the model can effectively learn the semantic information of unlabelled images, and are often implemented by data augmentation techniques. For example, Bortsova et al.\cite{r17} explored the equivariance of deformations in chest x-rays, which promoted consistent segmentation results between two different output branches. Huang et al.\cite{r18} used clipping and non-overlapping image patches as perturbations to measure the consistency of unlabelled images. In addition, the idea of "mixing" has gradually been introduced into consistency regularisation methods. For example, MixMatch \cite{r19} incorporates mixup as part of the perturbation process, applying perturbations to images along with traditional data augmentation methods. Basak et al.\cite{r20} apply Mixup only to images and also achieve good results in cardiac and abdominal MRI segmentation. Consistency regularisation methods usually have the advantage of being easy to implement, can be combined with other types of methods to further improve performance, and require adaptation of the perturbation methods based on the different objectives of the tasks. Using too many or inappropriate perturbation methods for the segmentation objectives can have a negative impact on model learning.


The second category is based on pseudo-labels in semi-supervised learning methods. This type of approach usually involves multiple models, where one model, based on learning from labeled images, generates pseudo-labels for unlabelled images, while another model uses these pseudo-labels for supervised learning on the unlabelled images. For example, in the Mean Teacher\cite{r12} pipeline, the student network is trained on the entire dataset, while the teacher network updates its parameters using the exponential moving average of the student network's parameters and generates pseudo-labels. Building on Mean Teacher, some research \cite{r21, r31, r46, r47} found that incorporating the consistency loss between regions of interest (ROIs) of the image, such as the ventricle and myocardium, and the Uncertainty-Aware Map or data augmentation as a regularisation term on top of Mean Teacher can improve the generalization of the model.

In addition, Peng et al.\cite{r15} introduced a pseudo-label-based semi-supervised algorithm called Deep Co-training (DCT) to ventricle segmentation research and demonstrated its effectiveness in this domain. Building on this, Zheng et al.\cite{r10} and Chen et al.\cite{r14} combined the uncertainty-aware map with DCT, resulting in improvements in segmentation accuracy for both ventricles and myocardium. Although there is still a gap compared to fully supervised learning, Chen's work showed that focal loss can address the issue of class imbalance in ventricle images. What's more, researches \cite{r48} and \cite{r49} prove that the distribution shift via adversarial perturbations also can improve the robustness of co-training. In addition, the CPS (Cross Pseudo Supervision) algorithm proposed by Chen et al.\cite{r11} involves the collaborative training of two models, each acting as a teacher network to generate pseudo-labels for the other, combined with CutMix for data augmentation. The process has similarities to knowledge distillation algorithms. Luo et al.\cite{r22} and Wang et al.  \cite{r50}, based on CPS, both demonstrate the feasibility of combining different types of networks, such as CNN\cite{r51} and ViT\cite{r45}. The semi-supervised algorithms come closest to the performance of fully supervised segmentation but require more computing resources and reasoning time.

In pseudo-label-based semi-supervised learning methods, the key to accurately learning semantic features from unlabelled images lies in the ability of the teacher network to generate reliable pseudo-labels. Currently, these methods often suffer from the problem that teacher networks rely too much on information from labeled images, making it difficult to further improve the reliability of pseudo-labels. Therefore, the key to improving the reliability of pseudo-labels is to enable the teacher network to effectively learn semantic features from the dataset.

\subsection{Data augmentation}


In machine learning models, a large amount of data is crucial for training highly accurate models. As deep learning models become more complex, they have a greater ability to learn from images, which can sometimes lead to overfitting. Overfitting occurs when a model performs well on the training set but fails to achieve the same level of accuracy on the test set. This phenomenon is primarily due to the model over-fitting the space of the training data, resulting in low generalization and poor performance on other sample sets within the same space. Overfitting occurs more frequently when the dataset is relatively small.

To address this problem, researchers introduce random perturbations to images before training, making it more difficult for the model to distinguish between samples and thus expanding the size of the dataset. Common augmentation techniques include rotations, flips, translations, stretches, sca-ling, color transformations, and the addition of Gaussian noi-se, which have shown good results in image classification tasks. To incorporate even more randomized augmentation data, Cubuk et al.  \cite{r23} introduced the AutoAugment algorithm. The algorithm integrates multiple augmentation techniques into a collection and uses a search model to find the most appropriate augmentation parameters for a deep learning network. These parameters are then used to determine the methods and magnitudes of augmentation in each case, resulting in richer and more effective augmentation data. Subsequently, Cubuk et al. proposed the RandAugment algorithm  \cite{r24}, which optimized the search process of AutoAugment by replacing subtask search with grid search. This not only reduced the computational complexity of the algorithm but also further improved the augmentation performance. Traditional data augmentation techniques have greatly reduced the data dependence of deep learning models and improved the performance in image classification tasks of these models. However, to maintain randomness during the augmentation process, these algorithms require additional computational resources.


French et al.\cite{r16} pointed out that in semantic segmentation, traditional augmentation techniques struggle to be effective due to the lack of low-density regions between classes in the data distribution. However, methods with higher diversity, such as Mixup and CutMix, have shown strong improvements in semantic segmentation. Mixup\cite{r25} is the first method to apply this idea to machine learning, mixing two images by adding them with weighted coefficients, resulting in better diversity. CutMix\cite{r26} combines Mixup with Cutout, where a rectangular region is cut out of one image and pasted into another. CowMix\cite{r27} improves on CutMix by introducing irregularly shaped regions for even more random mixing. There are now many methods that have applied the mixup concept to medical image analysis. For example, Wang et al.\cite{r28} introduced a TensorMixup method that extracts regions of interest from brain tumor images and mixes them using Mixup, resulting in a non-linear mixture that differs from traditional Mixup. Raj et al.\cite{r29} proposed a technique of cross-mixing, which differs from the single cropping of CutMix. This algorithm selects multiple block or striped areas from corresponding positions in two images, achieving high-density but distortion-free augmentation. It performs well on datasets relating to breast cancer, brain tumors, and more.

Data augmentation not only increases sample diversity but also highlights object features. ClassMix\cite{r30} extracts segmentation objects from one image and pastes them at the same location in a second image. This method creates new samples that expand the dataset while maintaining similarity to the original samples. It also increases the diversity of edge features of the segmentation objects. ClassMix shows good results on street scene segmentation datasets.

In summary, different data augmentation methods enable deep learning models to be more robust and to overcome problems with insufficient data. In addition, these methods can be further improved based on their objectives to enhance target features and drive the model to learn more semantic information.



\section{Methods}
\subsection{Problem formulation}
To make efficient use of unlabelled data, we introduce a novel data augmentation method called Scaling-up Mix with Multi-Class (SM$^2$C) in Meta Pseudo Labels. First, we formalize the problem of semi-supervised learning and semantic segmentation tasks. For the data set $D$, it comprises two components: $D=D_l\cup D_u$. The labeled dataset $D_l$ consists of images with manual annotations and can be represented as $D_l=(x_i,y_i)^{N_l}{i=1}$, where $N_l$ represents the amount of data contained in $D_l$. The samples in the unlabelled data set have no labels, which can be represented as $D_u=(x_j)^{N_u}{j=1}$, where $N_u$ is the amount of unlabelled data and $N_l\ll N_u$. We'll also define $f_T(\theta_T)$ as the teacher network and $f_S(\theta_S)$ as the student network. $\theta_T$ and $\theta_S$ are trainable parameters of the networks and are updated separately.

\subsection{Scaling-up Mix}
\begin{figure}[t] 
	\centering
	\includegraphics[width = .9\linewidth]{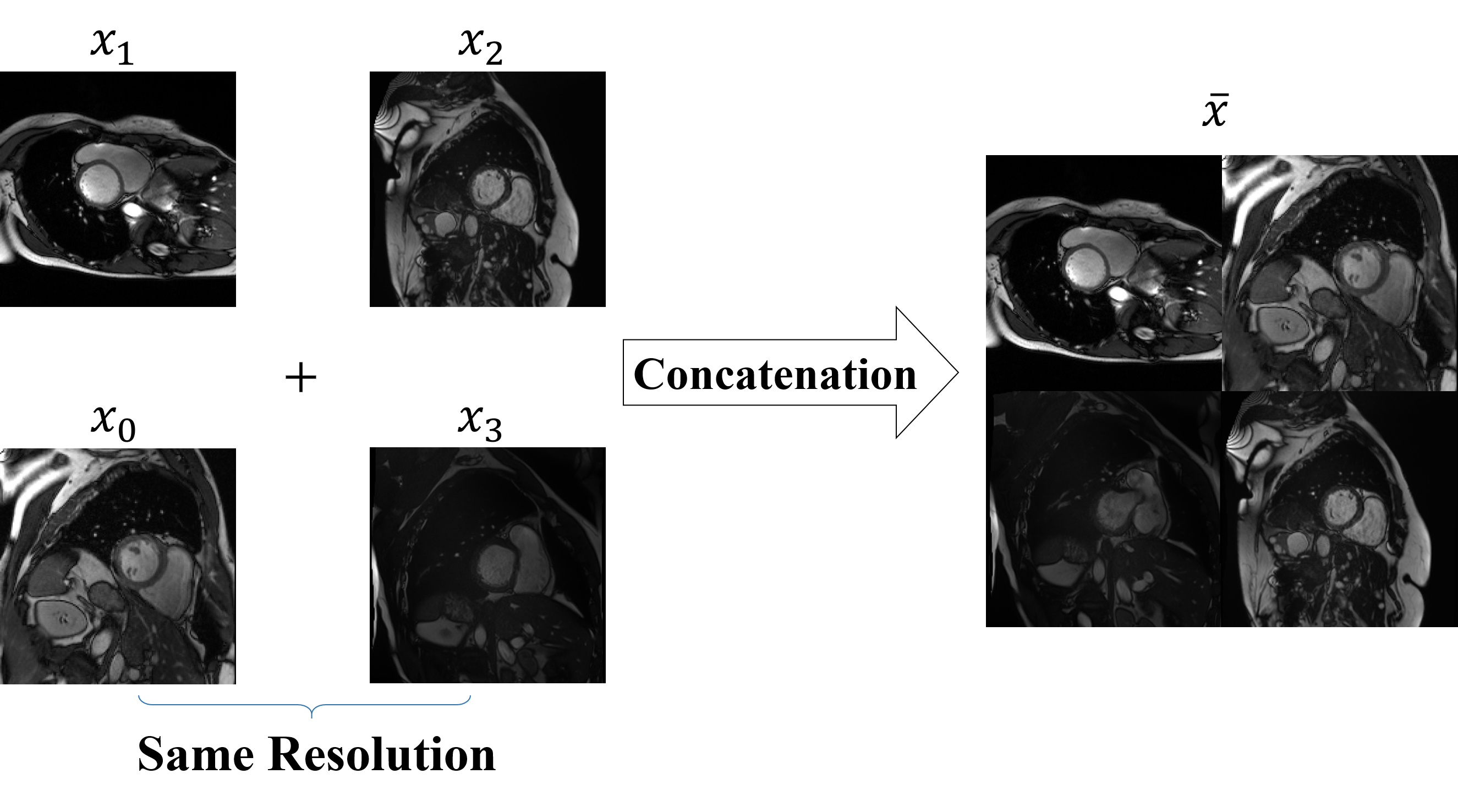}
        \caption{Illustration of Scaling-up Mix, an important part of SM$^2$C. Scaling-up Mix creates images with increased foreground-background diversity by concatenating four images.}
	\label{fig:fig2}
\end{figure}

\begin{figure*} 
	\centering
	\includegraphics[width = .9\linewidth]{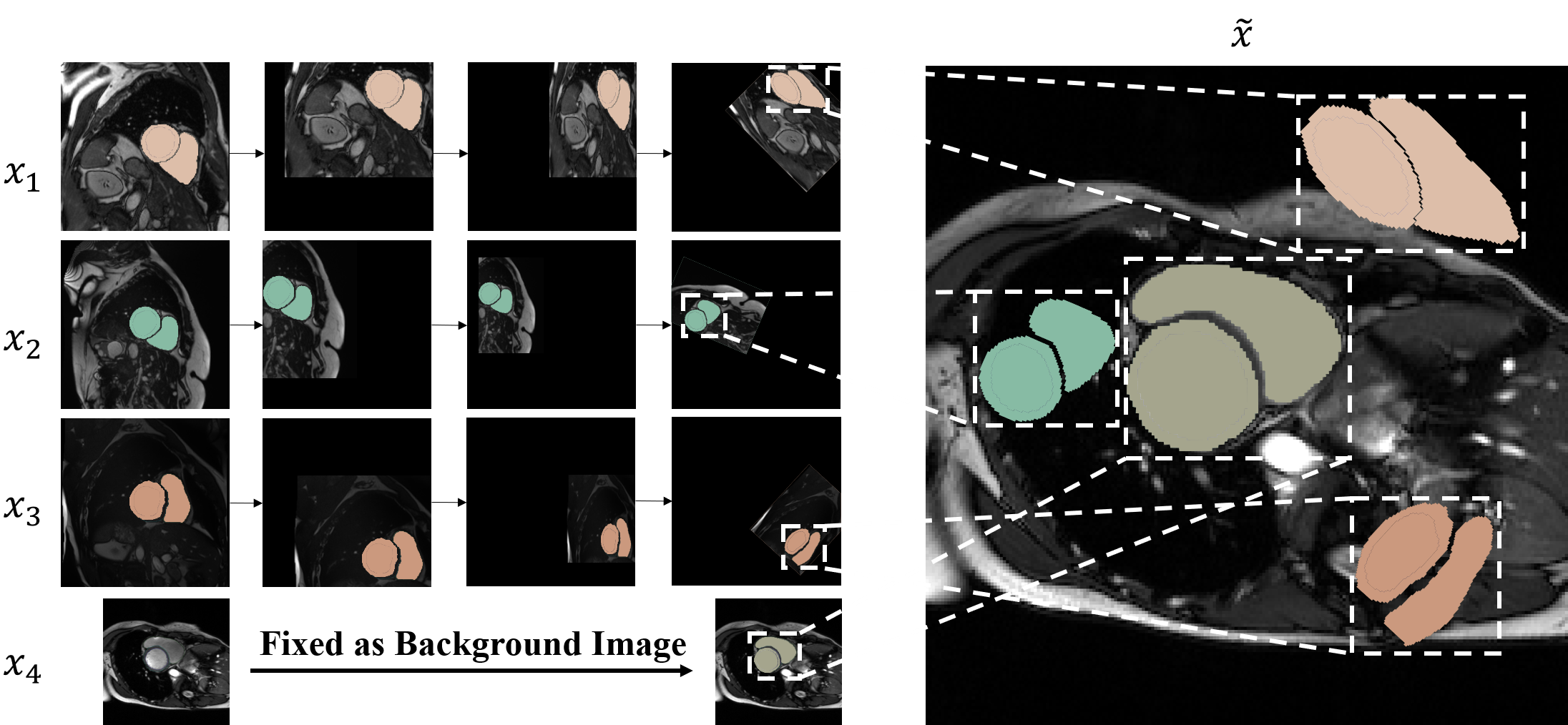}
        \caption{The Multi-Class-Jittering Mix operation first extracts the segmentation objects from each original image. After applying transformations such as deformation and translation to these objects, they are mixed into one of the original images. This process simulates the diversity of organ morphology as well as the diversity between foreground and background in medical images.}
	\label{fig:fig3}
\end{figure*}

\begin{figure*} 
	\centering
	\includegraphics[width = .9\linewidth]{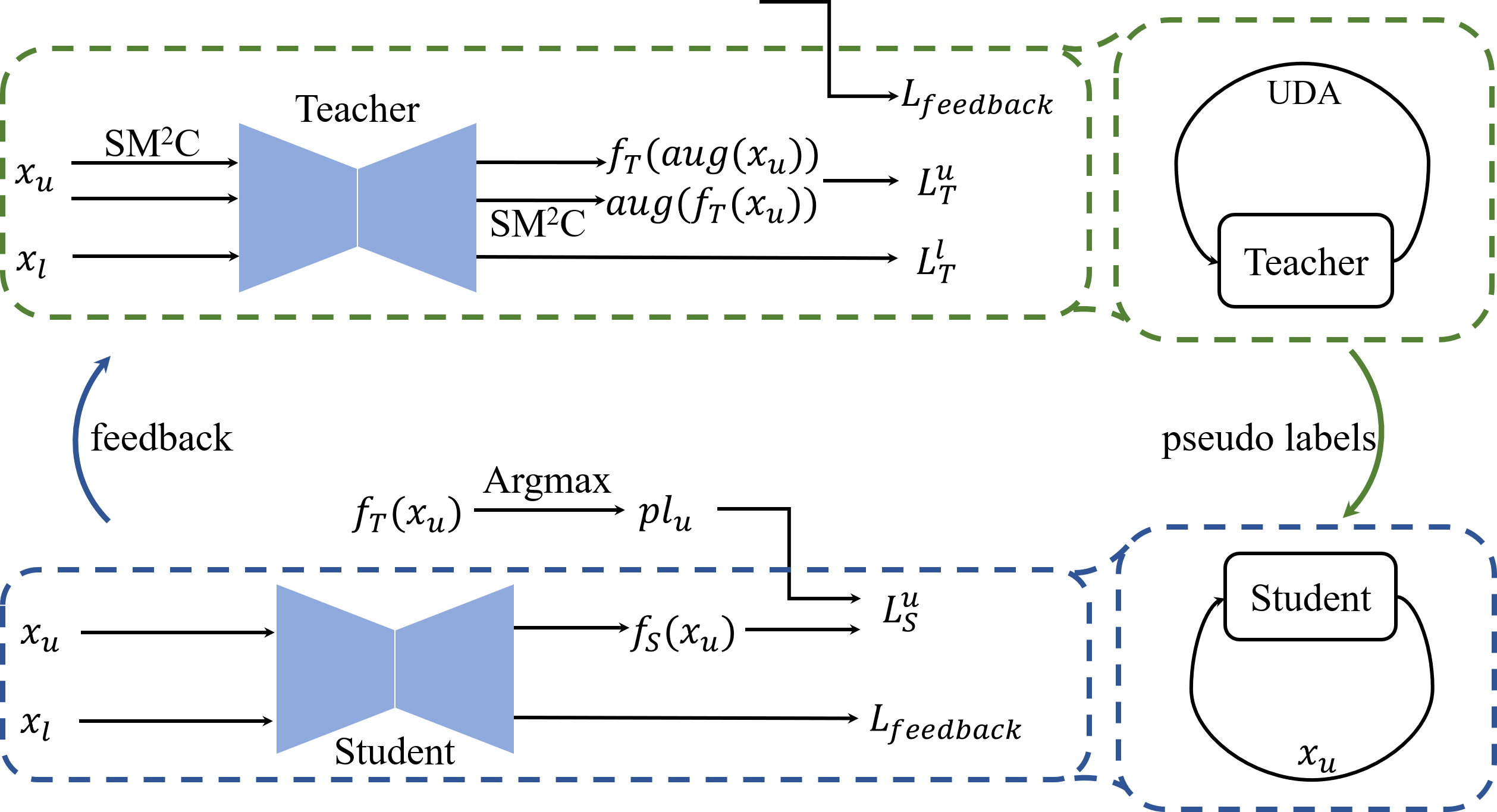}
        \caption{Illustration of the application of SM$^2$C in a semi-supervised framework based on MPL. The framework consists of two networks: the teacher network and the student network. The teacher network generates pseudo-labels for unlabelled images and the student network trains on these samples. In each iteration, the feedback generated by the student network guides the training of the teacher network. In addition, the UDA part, based on SM$^2$C, allows the teacher network to be trained on unlabelled images, allowing the teacher network to learn additional semantic information beyond the labeled images, thereby generating more reliable pseudo-labels.}
	\label{fig:fig4}
\end{figure*}

Empirical evidence has shown that training networks using multiple images in each iteration significantly contributes to improved network generalisation\cite{r40}. Among the parameters influencing training, the size of the mini-batch stands out as a key factor in regulating the number of samples during training. However, it's important to note that the effectiveness of using a larger mini-batch size is not guaranteed. Keskar et al.\cite{r39} demonstrated that under certain dataset configurations, larger batch sizes may have inferior performance compared to smaller batch sizes.

Beyond batch size adaptation, we argue that enriching the context of individual images has the potential to strength-en the network's ability to learn semantic features and mitigate overfitting. Our approach involves concatenating four different images to construct a novel input image characterized by increased size and more complex context, as visually illustrated in Figure \ref{fig:fig2}. The scaling procedure can be formalized as follows:
\begin{align}
\overline{x}={\rm Concat}(x_0,x_1,x_2,x_3)
\end{align}
Where $x_0$, $x_1$, $x_2$ and $x_3$ are 4 randomly selected images from the dataset $D_u$, resized to the same size before training, and we use $\rm Concat()$ to mix the images together. The image $\overline{x}$ is four times the size of the image $x_i\in D_u$. In particular, training the networks becomes more difficult because $\overline{x}$ contains 4 groups of segmentation objects, and the networks have to recognize more objects in a larger context, which avoids the possibility of missing objects of a certain category in an input image, as well as increasing the diversity of images in model training.

\subsection{Multi-Class Mix}
French et al.\cite{r16} emphasized the need for robust and diverse augmentation techniques in image segmentation instead of conventional methods designed for classification ta-sks. In particular, mixing-based approaches such as CutMix, Cow-Mix, and FMix have demonstrated their effective-ness in image segmentation.

However, the context is different for medical images, whe-re segmentation accuracy along boundaries is essential to meet the demands of clinical diagnosis. Methods such as CutMix, which mixes image patches, can compromise the integrity of object contours. In addition, objects of some specific categories are not always included in MRI or CT, which can lead to some degree of inter-class imbalance. To strengthen the focus of the network on contours, we adopt a strategy of cropping objects from one image and integrating them into another. Let $x_0$, $x_1$, $x_2$ and $x_3$ be the components of a mixed image, where each $x_i$ represents a single image. The augmentation procedure can be described by the following equations:
\begin{gather}
M=\sum^{K}_{k\neq i}M_k\\
\tilde{x}_i=(1-M)\odot x_i+\sum^{K}_{k\neq i} (M_k\odot x_{k})
\end{gather}
$K$ denotes the number of images engaged in the mixing process. $M$ is a binary mask derived from objects cropped out of images except $x_i$. $M_k$ is generated from annotations or predictions of $x_k$, with pixel values of 1 indicating object regions and 0 otherwise. We randomly select various categories and employ dot product $\odot$ to extract and paste objects onto $x_i$, augmenting the number of segmentation objects within a single image. This operation is executed across the remaining three images before scaling-up.

Unlike other types of mixing algorithms, Multi-Class Mix preserves the structural integrity of segmentation object contours by increasing the number of objects within input images and enhancing the diversity of the background surrounding the objects. This approach provides the networks with a more comprehensive understanding of the features of segmentation objects in different backgrounds.

\subsection{Multi-Class-Jittering Mix}
In the process of MRI, the objects in different images usually have significant differences in morphology. The segmentation network requires the ability to identify objects in the same category with different shapes. To this end, we introduce deformation operations in the proposed method. The operations include translation, rotation, flipping, and scaling et al and are implemented by affine matrix $A$, which can be represented as follows:

In the context of MRI, the objects present in distinct images often exhibit substantial variations in morphology. For an effective segmentation network, the capacity to discern objects of the same category but with dissimilar shapes is imperative. In pursuit of this goal, we introduce deformation operations within our proposed method. These operations encompass translation, rotation, flipping, scaling, and more. They are effectuated using an affine matrix denoted as $A$, which can be succinctly expressed as:
\begin{gather}
M_i^{'}=A(s,r,f,\theta)\cdot M_i
\end{gather}
where $s$, $\theta$, $f$, $t$, and $r$ represent the factors to adjust size, flipping, rotation, translation, and length-width of objects. $M_i$ contains the objects in image $x_i$. Rotation and flipping can prompt the segmentation network to adapt a wide range of medical datasets. Size, translation, and length-width adjustment can simulate the morphological differences caused by different patients and instruments. Figure \ref{fig:fig3} illustrates the pipeline of how Multi-Class-Jittering Mix works.

The Scaling-up Mix with Multi-Class is to mix the segmentation objects into multiple images after performing deformation operations on the objects, before performing Sca-ling-up Mix to expand the size of input images, as shown in Figure \ref{fig:fig1}. The overall augmentation process of SM$^2$C is shown in Algorithm 1

\subsection{Semi-Supervised Framework}
Inspired by the MPL framework, our semi-supervised algorithm is based on the pseudo-label framework. It utilizes feedback from the student network to effectively adjust the training direction of the teacher network. Additionally, the algorithm benefits from the SM$^2$C method to enhance the learning of semantic information related to segmentation objects in medical images. As illustrated in Figure \ref{fig:fig4}, the student network employs pseudo-labels for supervised learning on unlabeled images. The training process involves the following loss functions:
\begin{align}
\label{equ:equ5}
L_S^u=\mathcal{L}_{ce}(f_S(x_u;\theta_S^{(t)}),f_T(x_u;\theta_T))
\end{align}
Where $\theta_S^{(t)}$ represents the parameters of the student network at the $t$-th iteration and $pl_u$ stands for the pseudo-label generated by the teacher network for the unlabeled image $x_u$. Using the loss $L_S^u$, the parameters of the student network can be updated using the following equation:
\begin{align}
\label{equ:equ6}
\theta_S^{(t+1)}=\theta_S^{(t)}-\eta_S\cdot\bigtriangledown_{\theta_S^{(t)}} L_S^u
\end{align}
Additionally, the teacher network is trained synchronously with the student network, and its training process involves three components in the loss function:
\begin{equation}
L_T=L_T^l+L_T^u+L_T^{feedback}
\end{equation}
Where $L_T^l$ is obtained by the teacher network through supervised learning on labeled images and can be expressed as:
\begin{equation}
L_T^l=\mathcal{L}_{ce}(y_l,f_T(x_l;\theta_T^{(t)}))
\end{equation}
Where $\theta_T^{(t)}$ represents the parameters of the teacher network at the $t$-th iteration, and $y_l$ denotes the ground truth annotation of the image $x_l$. To further leverage the unlabeled images and enable the teacher network to better learn the semantic information from them, we introduce the Unsupervised Data Augmentation (UDA) method to learn from the unlabeled images. Additionally, we use SM$^2$C as the perturbation method. The loss $L_T^u$ for this process can be expressed as:
\begin{equation}
L_T^u=\omega \cdot \mathcal{L}_{cons}(f_T(aug(x_u);\theta_T^{(t)}),aug(f_T(x_u;\theta_T^{(t)})))
\end{equation}
Where $\mathcal{L}_{cons}$ is a consistency loss function, $\omega$ represents the weight of the consistency loss in the overall loss, and $aug()$ represents the data augmentation method SM$^2$C. In the computation of $\mathcal{L}_{cons}$, it's essential to ensure that the unlabeled image $x_u$ and its pseudo label $f_T(x_u;\theta_T^{(t)})$ both involved in the computation have undergone the same augmentation transformations.

From Equation~\ref{equ:equ5} and Equation~\ref{equ:equ6}, it's evident that the student network loss $L_S^l$ is a function of the parameters $\theta_S$, which change according to the variations in the teacher network parameters $\theta_T$. Consequently, the teacher network can update its parameters based on the student network's loss $L_S^l$ on labeled images. This update process is defined as:
\begin{align}
L_T^{feedback} &=L_S^l(\theta_S^{(t+1)})\\
&=L_S^l(\theta_S^{(t)}-\bigtriangledown_{\theta_S^{(t)}}L_S^u(\theta_S^{(t)},\theta_T^{(t)}))
\end{align}
The feedback $L_T^{feedback}$ is represented by the loss $L_S^l$. Throu-gh this equation, the teacher network can obtain the gradient of the updatable parameters $\theta_T$ through backpropagation, thereby achieving parameter updates based on the performance of the student network.
\begin{algorithm}
	\renewcommand{\algorithmicrequire}{\textbf{Input:}}
	\renewcommand{\algorithmicensure}{\textbf{Output:}}
	\caption{Scaling-up Mix with Multi-Class}
	\label{alg1}
	\begin{algorithmic}[1]
            \REQUIRE Segmentation network$f(\theta)$, labeled dataset $D_l$, unlabeled datase$D_u$, number of classes $C$, number of mixed images $K$.

            \STATE Enumerate$\lbrace x_u^0,x_u^1,...,x_u^K\rbrace \in D_u$
            \STATE $n \leftarrow 0$
            \REPEAT 
            \STATE $p_n=argmax~f(x_u^n;\theta)$
            \STATE Randomly select $c$ categories, $c < C$
            \STATE For all i, j: $M_n$ = $\begin{cases}
		1, if \ p_n(i,j)\in c\\
		0, otherwise\\
		\end{cases}$\\
            \UNTIL $n$ equals $K$

            \STATE $n \leftarrow 0$
            \REPEAT
            \STATE $M=Jitter(\sum^{K}_{k\neq n}M_k)$
		\STATE $\tilde{x}_n=(1-M)\odot x_n+\sum^{K}_{k\neq n} (M_k\odot p_{k})$
		\STATE $\tilde{p}_n=(1-M)\odot p_n+\sum^{K}_{k\neq n} (M_k\odot p_{k})$
            \UNTIL $n$ equals $K$
            
            \STATE $\overline{x}=Concat(\tilde{x}_0,\tilde{x}_1,...,\tilde{x}_k)$
            \STATE $\overline{p}=Concat(\tilde{p}_0,\tilde{p}_1,...,\tilde{p}_k)$
            
		\ENSURE $\overline{x}$, $\overline{p}$.
	\end{algorithmic}  
\end{algorithm}

\section{Experiments and results}
\subsection{Dataset}
\begin{itemize} 
\item \textbf{ACDC dataset}: The ACDC benchmark dataset, as introduced in \cite{r32}, comprises a collection of 200 images sourced from short-axis cardiac cine-MRI scans from 100 patients. Each patient is represented by a pair of images corresponding to the end-diastolic (ED) and end-systolic (ES) cardiac phases. Image resolutions span from $0.70 \times 0.70$ mm to $1.92 \times 1.92$ mm in-plane, while through-plane resolutions vary between 5 mm and 10 mm. Clinical experts have annotated all 200 images, yielding segmentation masks delineating regions of significance, including the left ventricle (LV), right ventricle (RV), and left ventricle myocardium (Myo). For our experiments, we divide the 200 images into the training set, validation set, and test set, with 140, 20, and 40 samples respectively. Given the large inter-slice spacing present in the data, we conduct 2D segmentation rather than 3D segmentation\cite{r34}.
\item \textbf{SCGM dataset}: The dataset utilized for the multi-center, multi-source spinal cord grey matter segmentation challenge (SCGM)\cite{r33}, focuses on the identification and segmentation of grey and white matter within spinal MRI images. These images originate from four distinct medical centers, each contributing 20 samples. Within each center's set, 10 images are annotated by experienced experts. Significantly, the images included in this dataset vary in terms of their acquisition details, stemming from disparate MRI systems and diverse imaging parameters. This variation results in a notable diversity between samples from distinct sources. The image resolution spans from $0.25 \linebreak \times 0.25 \times 0.25 \text{ mm}$ to $0.5 \times 0.5 \times 0.5 \text{ mm}$, as well as the number of axial slices, ranging from 3 to 28.  We split the dataset into labeled (4 patients), unlabeled (28 patients), validation image (4 patients), and test image subsets (4 patients). We also conduct 2D segmentation in SCGM, which contains 57 labeled images and 383 unlabeled images.
\item \textbf{Prostate dataset}: Prostate dataset is one of the ten sub-tasks of the Medical Segmentation Decathlon Cha-llenge\cite{r35}. The dataset includes 48 multi-parametric MRIs (only 32 were given with ground truth). The ground truth of the whole prostate consists of transverse T2-weighted scans with resolution 0.6 x 0.6 x 4 mm and the apparent diffusion coefficient (ADC) map (2 x 2 x 4 mm). We split the dataset into labeled (4 patients), unlabeled (24 patients), and test image subsets( 4 patients).
\end{itemize}

To ensure a fair comparison, referring to the data preprocessing methods in previous research work, we normalize the value of image pixels to [0,1] and resize all images to $256\times 256$ before the training phase. We perform some standard data augmentation operations on images, including random rotation within the range of $\pm 20^{\circ}$ and random flipping.

\subsection{Implementation details and evaluation metrics}
For our dual-network framework, we set U-Net as the segmentation network both for the teacher network and the student network. Both of them are in the same architecture, with the poly learning rate strategy, where the initial learning rate was set to 0.01. The pre-trained loading model of U-Net was trained on labeled samples of each medical image dataset. The total iterations are 30k and the Batch size for labeled and unlabeled samples are both set to 6. Our experiments are conducted by using PyTorch on a server with 4 NVIDIA 3090 GPUs (24GB). 

The evaluation metrics used in our experiments are the Dice coefficient and Hausdorff distance. The dice coefficient (DSC) is to measure the similarity between predictions P and ground-truth G:
\begin{align}
DSC=\frac{2\cdot |P\cap G|}{|P|+|G|}
\end{align}
Besides, Hausdorff distance (HD) is a measure of the boundary distance between the prediction P and ground-truth G by calculating the maximum distance of the two nearest points p and g :
\begin{align}
{HD}(P,G)=max\lbrace {\rm dist}(g,p), {\rm dist}(p,g)\rbrace
\end{align}
The average and standard deviation of the DSC score and Hausdorff distance over all patients are reported.


\begin{figure*} 
	\centering
	\includegraphics[width = .9\linewidth]{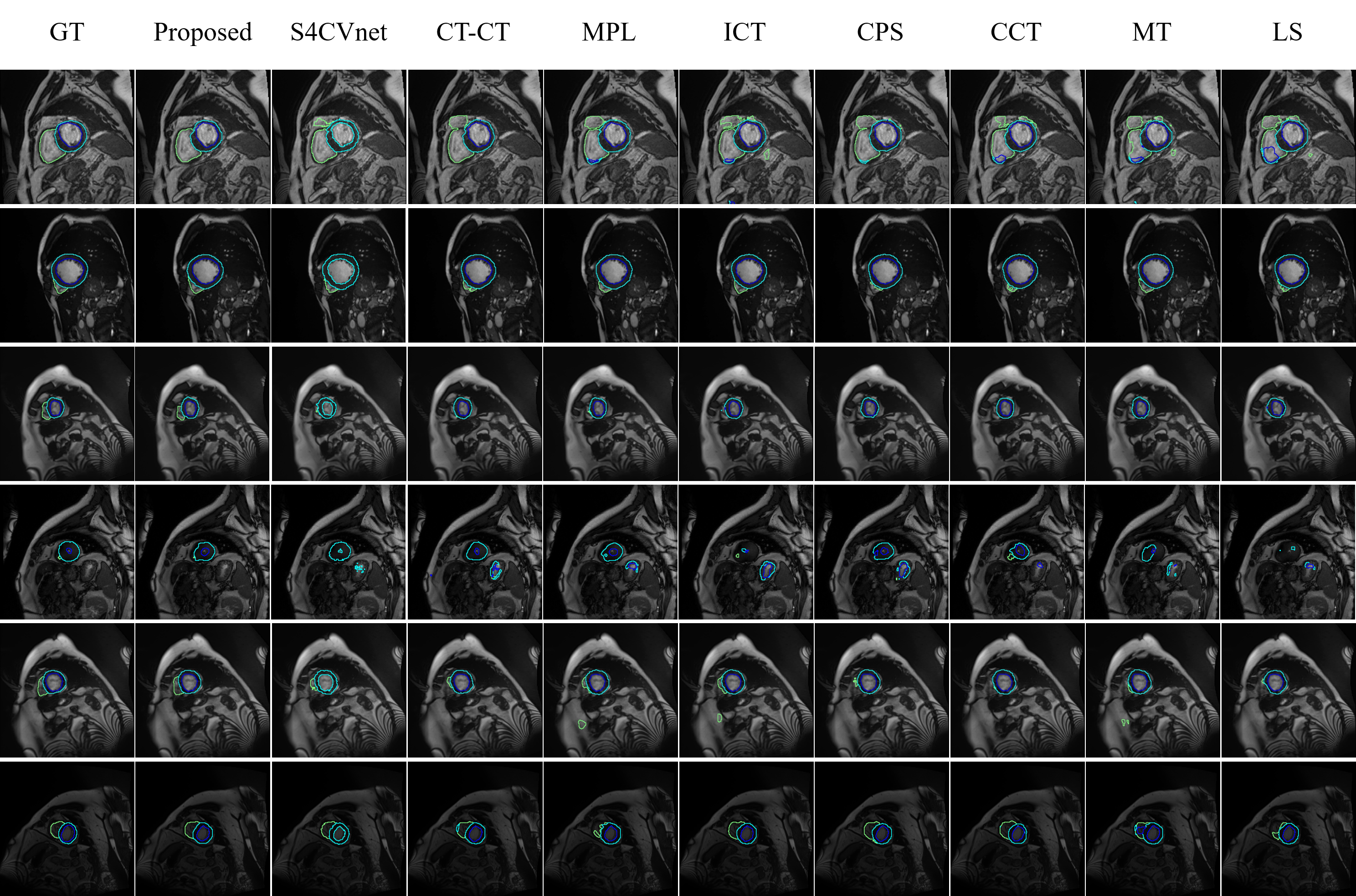}
	\caption{Examples of segmentation results for the ACDC dataset with 10\% labeled training examples. From left to right: Ground Truth (GT), the proposed framework, S4CVnet\cite{r50}, Cross Teaching between CNN Transformers\cite{r36} (CT-CT), Meta Pseudo Label (MPL)\cite{r15}, Interpolation Consistency Training\cite{r37} (ICT), Cross Pseudo Supervision\cite{r11} (CPS), Cross Consistency Training\cite{r36} (CCT), Mean Teacher\cite{r12} (MT), and Limited Supervised (LS) Learning.}
	\label{fig:fig5}
\end{figure*}

\begin{figure*} 
	\centering
	\includegraphics[width = .9\linewidth]{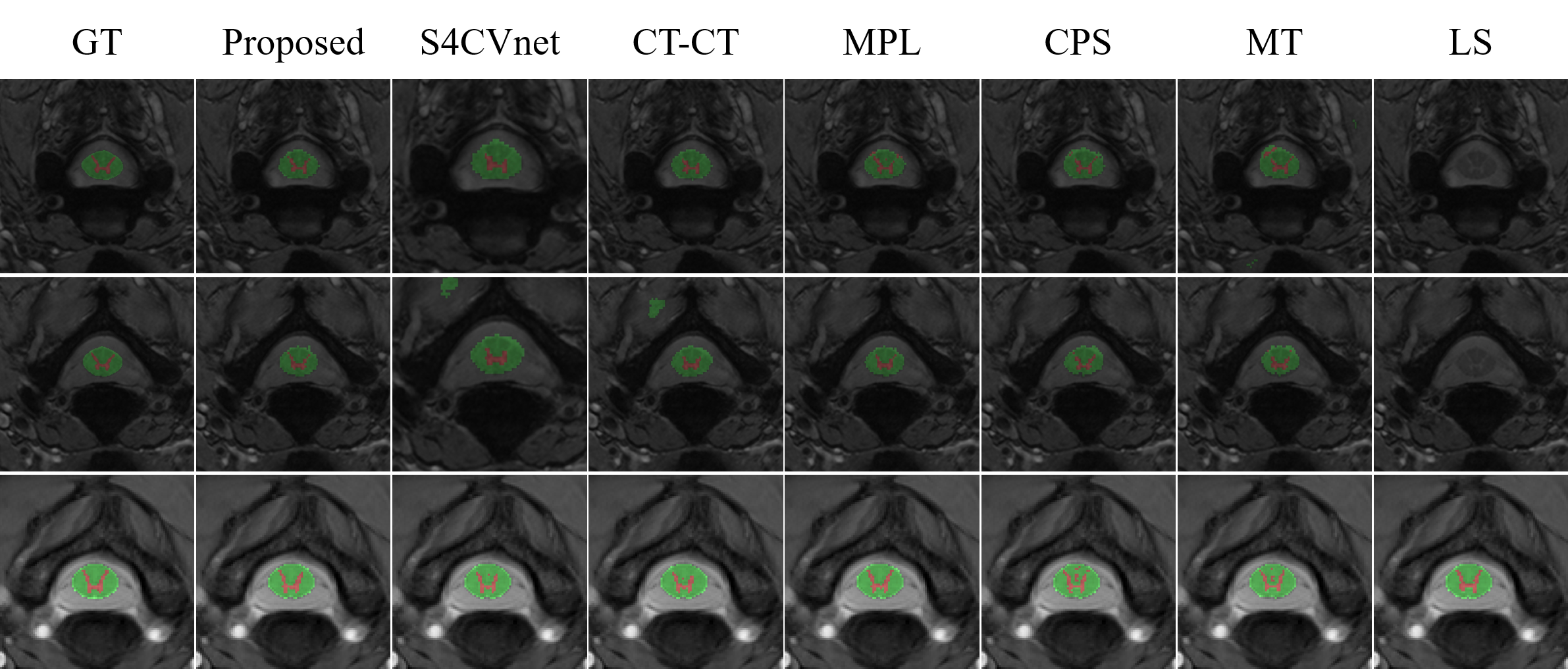}
	\caption{Examples of segmentation results for the SCGM dataset with 1/8 of the labeled training examples. From left to right: Ground Truth (GT), the proposed framework, S4CVnet\cite{r50}, Cross Teaching between CNN Transformers\cite{r36} (CT-CT), Meta Pseudo Label (MPL), Interpolation Consistency Training\cite{r37} (ICT), Cross Pseudo Supervision\cite{r11} (CPS), Cross Consistency Training\cite{r36} (CCT), Mean Teacher\cite{r12} (MT), and Limited Supervised (LS) Learning.}
	\label{fig:fig6}
\end{figure*}

\begin{figure*} 
	\centering
	\includegraphics[width = .9\linewidth]{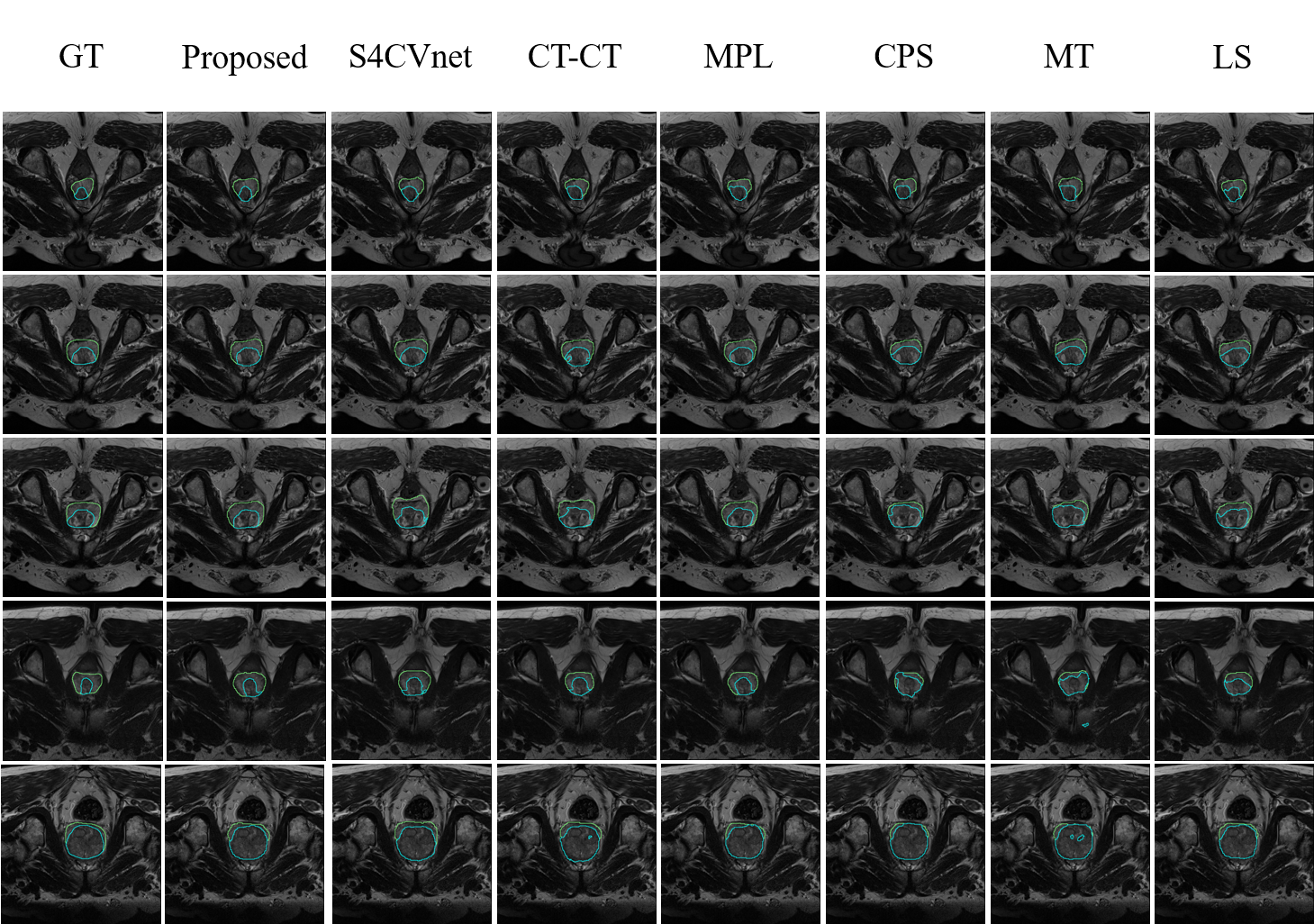}
	\caption{Examples of segmentation results for the prostate dataset with 1/7 of the labeled training examples. From left to right: Ground Truth (GT), the proposed framework, S4CVnet\cite{r50}, Cross Teaching between CNN Transformers\cite{r36} (CT-CT), Meta Pseudo Label (MPL), Interpolation Consistency Training\cite{r37} (ICT), Cross Pseudo Supervision\cite{r11} (CPS), Cross Consistency Training\cite{r36} (CCT), Mean Teacher\cite{r12} (MT), and Limited Supervised (LS) Learning.}
	\label{fig:fig7}
\end{figure*}

\subsection{Comparison existing works}

\subsubsection{ACDC dataset}

Table \protect\ref{tab:table1} shows the results of eight semi-supervised algorithms applied to the cardiac dataset. In the context of 7 cases, our proposed method demonstrates a notable performance advantage, outperforming other approaches by approximately 3\% improvement in mean DSC and achieving a 3mm reduction in mean HD. Notably, our method also excels in both regional and contour detection when segmenting ventricles and myocardium, respectively. What's more, the variance in brackets represents the stability of the predictions in the test dataset. Compared to SOTA, the proposed method has a lower variance, which means that the method can stably produce highly accurate predictions for each sample in the inference stage. Moreover, the proposed method significantly outperforms the S4CVnet\cite{r50} in the setting of 3 cases, showing a remarkable 15\% improvement in the DSC similarity coefficient (DSC). However, our method shows higher HD values for the left ventricle and myocardium. We suggest that this may be due to their intricate shapes, which span the bottom-to-top dimensions of the heart in the images, and highly augmented samples by SM$^2$C may hinder the learning of the teacher model to some extent compared to S4CVnet. Conversely, our method shows a superior ability to detect right ventricles, resulting in lower HD values compared to the CT-CT approach. MPL does not show significant competitiveness In both settings compared to other methods without considering consistency loss.

Figure \ref{fig:fig5} visually contrasts the segmentation results of different semi-supervised frameworks, using 7 labeled cases for training. Our method outperforms S4CVnet by achieving more intact object segmentations. In particular, our approach excels in the accurate segmentation of right ventricles, closely resembling the ground truth. In addition, our method has fewer false positive regions, highlighting its accuracy.

\begin{table*}[width=1.9\linewidth,cols=10,pos=h]
  \centering
  \setlength{\tabcolsep}{4pt}
  \fontsize{7}{10.5}\selectfont
  \caption{Comparison between the proposed method and other methods on the ACDC dataset.}\label{tab:table1}
    \begin{tabular}{clcccccccc}
    \toprule
    \multicolumn{2}{c}{\multirow{2}{*}{Methods}}&
    \multicolumn{2}{c}{RV}&\multicolumn{2}{c}{Myo}&\multicolumn{2}{c}{LV}&\multicolumn{2}{c}{Mean}\cr
    \cmidrule(lr){3-10}
    &&{$\rm DSC(\%)$}&{$\rm HD_{95}(mm)$} &{$\rm DSC(\%)$}&{$\rm HD_{95}(mm)$} &{$\rm DSC(\%)$}&{$\rm HD_{95}(mm)$} &{$\rm DSC(\%)$}&{$\rm HD_{95}(mm)$}\cr
    \midrule
    \multirow{9}{*}{3 cases}
    &LS&37.70(0.23)&50.15(7.89)&51.49(0.07)&14.28(9.84)&61.02(0.11)&19.01(17.28)&50.07(0.14)&27.81(11.67)\cr
    &CCT&40.8(0.32)&34.2(23.6)&64.7(0.21)&22.4(19.4)&70.4(0.24)&27.1(23.9)&58.6(0.26)&27.9(22.3)\cr
    &CPS&43.8(0.31)&35.8(24.1)&65.2(0.21)&18.3(16.3)&72(0.25)&22.2(22.0)&60.3(0.26)&25.5(20.8)\cr
    &ICT&44.8(0.33)&23.8(16.4)&62(0.24)&20.4(20.4)&67.3(0.29)&24.1(21.1)&58.1(0.28)&22.8(19.3)\cr
    &MT&40.3(0.29)&53.9(28.7)&58.6(0.25)&23.1(20.5)&70.9(0.25)&26.3(25.8)&56.6(0.26)&34.5(25.0)\cr
    &MPL&48.18(0.32)&21.06(22.76)&62.24(0.28)&11.51(17.4)&69.19(0.3)&15.13(21.89)&59.87(0.3)&15.90(20.69)\cr
    &CT-CT&67.7(0.29)&10.04(19.51)&65.66(0.26)&3.90(5.11)&73.96(0.28)&5.60(11.04)&69.11(0.28)&6.51(11.89)\cr
    &S4CVnet&69.12(0.26)&5.52(14.84)&68.23(0.24)&{\bf 1.37(1.27)}&72.58(0.30)&{\bf 3.17(10.77)}&69.98(0.27)&{\bf 3.35(8.96)}\cr
    &Proposed&{\bf 84.69(0.08)}&{\bf 4.22(7.57)}&{\bf 82.96(0.05)}&3.41(5.29)&{\bf 87.93(0.10)}&6.79(10.20)&{\bf 85.19(0.08)}&4.80(7.69)\cr
    \cmidrule(lr){2-10}
    \cmidrule(lr){1-10}
    \multirow{9}{*}{7 cases}
    &LS&73.37(0.18)&6.91(16.91)&80.64(0.06)&4.25(10.10)&87.60(0.09)&7.35(14.10)&80.54(0.11)&6.17(13.71)\cr
    &CCT&83.65(0.09)&4.37(8.11)&82.68(0.05)&5.65(10.9)&89.8(0.09)&8.20(14.53)&85.38(0.08)&6.07(11.18)\cr
    &CPS&82.86(0.13)&2.75(3.95)&82.39(0.05)&6.77(11.48)&87.13(0.11)&13.56(17.66)&84.13(0.1)&7.69(11.03)\cr
    &ICT&82.01(0.15)&4.22(7.19)&81.73(0.05)&9.91(15.05)&87.03(0.1)&16.07(21.34)&83.59(0.1)&10.07(14.53)\cr
    &MT&74.15(0.21)&4.83(6.6)&79.47(0.07)&9.54(15.44)&84.68(0.13)&14.16(18.64)&79.43(0.14)&9.51(13.56)\cr
    &MPL&83.47(0.1)&3.86(8.44)&82.17(0.05)&6.10(11.94)&88.69(0.09)&7.95(15.82)&84.77(0.08)&5.97(12.07)\cr
    &CT-CT&86.09(0.09)&2.72(3.75)&83.97(0.04)&5.78(10.6)&89.8(0.09)&8.10(14.84)&86.62(0.07)&5.53(9.73)\cr
    &S4CVnet&86.04(0.09)&3.51(8.15)&84.45(0.04)&4.21(9.20)&90.79(0.08)&5.8(10.63)&87.1(0.07)&4.51(9.33)\cr
    &Proposed&{\bf 90.19(0.06)}&{\bf 1.29(0.83)}&{\bf 87.33(0.04)}&{\bf 1.19(0.31)}&{\bf 92.93(0.06)}&{\bf 2.39(4.80)}&{\bf 90.15(0.05)}&{\bf 1.62(1.98)}\cr
    \cmidrule(lr){1-10}
    &FS&90.81(0.06)&1.67(3.33)&89.46(0.03)&1.36(2.19)&94.93(0.03)&1.87(5.3)&91.73(0.04)&1.63(3.61) \cr
    \bottomrule
    \end{tabular}
\end{table*}

\subsubsection{SCGM dataset}
To validate the effectiveness of our proposed method on different medical images, we evaluated its performance on MRI images of the grey matter of the spinal cord. As shown in Table \ref{tab:table2}, many semi-supervised frameworks excel at detecting white matter over grey matter. This is due to the regular shape of white matter, which makes it easier to distinguish from the background. In particular, our method stands out in this evaluation, outperforming other semi-supervised approaches. It achieves higher DSC values for both grey and white matter, 82.89\% and 90.22\% respectively. In terms of contour accuracy, our method yields lower mean Hausdorff distance values (grey matter: 1.10, white matter: 1.18), demonstrating its exceptional ability to delineate contours for both grey and white matter. What's more, we can also observe that the proposed method remains highly accurate in the test dataset due to a lower variance.

As shown in Figure~\ref{fig:fig6}, due to the significant difference in area between grey and white matter in this dataset, there is a degree of imbalance between classes. In addition, the shape of grey matter is more complex, often leading to cases of missegmentation or under-segmentation. For example, the dorsal horns in the third image segmented by CT-CT are shorter than the ground truth. Overall, the proposed algorithm performs better than other algorithms in terms of fine detail recognition in grey matter. It provides a more complete segmentation of the posterior horn. Furthermore, cases of mis-segmentation in grey matter are relatively rare, indicating the superiority of the proposed algorithm in distinguishing between grey matter, white matter and background.

\begin{table*}[t]
  \centering
  \setlength{\tabcolsep}{6pt}
  \fontsize{7.5}{10.5}\selectfont
  \caption{Comparison between the proposed method and other methods on the SCGM dataset.}\label{tab:table2}
    \begin{tabular}{lcccccc}
    \toprule[1pt]
    \multirow{2}{*}{Method}&
    \multicolumn{2}{c}{Grey matter}&\multicolumn{2}{c}{White matter}&\multicolumn{2}{c}{Mean}\cr
    \cmidrule(lr){2-7}
    &{$\rm DSC(\%)$}&{$\rm HD_{95}(mm)$} &{$\rm DSC(\%)$}&{$\rm HD_{95}(mm)$} &{$\rm DSC(\%)$}&{$\rm HD_{95}(mm)$}\cr
    \midrule
    LS&60.126(0.35)&13.35(21.16)&67.726(0.39)&53.75(91.36)&63.926(0.37)&33.55(56.26)\cr
    CPS&78.09(0.03)&1.87(0.71)&88.23(0.04)&1.64(0.89)&83.16(0.03)&1.76(0.8)\cr
    MT&78.87(0.05)&1.97(1.69)&87.29(0.05)&22.14(36.38)&83.08(0.05)&12.06(19.03)\cr
    MPL&82.17(0.05)&1.18(0.32)&89.36(0.05)&1.31(0.54)&85.77(0.05)&1.25(0.43)\cr
    CT-CT&80.44(0.04)&1.33(0.41)&89.21(0.04)&14.28(22.76)&84.82(0.04)&7.80(11.59)\cr
    S4CVnet&79.85(0.05)&1.85(1.03)&89.19(0.05)&1.50(0.87)&84.52(0.05)&1.67(0.95)\cr
    Proposed&{\bf 82.55(0.05)}&{\bf 1.10(0.18)}&{\bf 89.68(0.05)}&{\bf 1.25(0.43)}&{\bf 86.12(0.05)}&{\bf 1.18(0.31)}\cr
    \midrule
    FS&82.64(0.04)&1.10(0.18)&91.34(0.03)&1.21(0.21)&86.99(0.04)&1.16(0.19) \cr
    \bottomrule[1pt]
    \end{tabular}
\end{table*}

\subsubsection{Prostate dataset}
We validated the effectiveness of our framework in segmenting the central gland and the peripheral zone in images from the prostate dataset. Our experiments were performed on 2D MRI scan slices resized to 256 × 256 pixels. The results are summarised in table \ref{tab:table7}. Our method showed improved stability (lower standard deviation) leading to improved accuracy, outperforming limited supervised learning. While not achieving full supervised baseline performance, our approach improved accuracy from 59.61\% to 67.20\% and reduced HD from 12.68 mm to 10.29 mm for CT-CT. Figure~\ref{fig:fig7} shows visual examples of the segmentation results achieved by our methods.

\begin{table*}[t]
  \centering
  \setlength{\tabcolsep}{6pt}
  \fontsize{7.5}{10.5}\selectfont
  \caption{Comparison between the proposed method and other methods on the Prostate dataset.}\label{tab:table7}
    \begin{tabular}{lcccccc}
    \toprule[1pt]
    \multirow{2}{*}{Method}&
    \multicolumn{2}{c}{Peripheral zone
}&\multicolumn{2}{c}{Central gland
}&\multicolumn{2}{c}{Mean}\cr
    \cmidrule(lr){2-7}
    &{$\rm DSC(\%)$}&{$\rm HD_{95}(mm)$} &{$\rm DSC(\%)$}&{$\rm HD_{95}(mm)$} &{$\rm DSC(\%)$}&{$\rm HD_{95}(mm)$}\cr
    \midrule
    LS&47.64(0.17)&10.72(5.14)&65.56(0.18)&13.57(10.37)&56.60(0.18)&12.14(7.75)\cr
    CPS&37.34(0.19)&13.59(8.53)&63.16(0.20)&15.09(13.13)&50.25(0.19)&14.34(10.83)\cr
    MT&42.16(0.17)&10.61(3.44)&54.53(0.20)&15.92(14.69)&48.34(0.19)&13.27(9.07)\cr
    MPL&53.75(0.16)&8.79(5.29)&72.18(0.21)&10.38(9.20)&62.97(0.18)&9.58(7.25)\cr
    CT-CT&49.58(0.21)&15.89(16.86)&69.65(0.21)&9.46(7.92)&59.61(0.21)&12.68(12.39)\cr
    S4CVnet&47.73(0.16)&10.93(4.53)&71.58(0.21)&{\bf 9.22(8.17)}&59.66(0.18)&10.07(6.35)\cr
    Proposed&{\bf 61.69(0.10)}&{\bf 6.99(2.83)}&{\bf 72.71(0.18)}&13.59(10.07)&{\bf 67.20(0.14)}&{\bf 10.29(6.45)}\cr
    \midrule
    FS&64.88(0.07)&6.60(4.60)&79.79(0.10)&4.71(1.61)&72.33(0.08)&5.65(3.10 ) \cr
    \bottomrule[1pt]
    \end{tabular}
\end{table*}

\subsection{Ablation study}


First, we examine the contributions of the different components of the SM$^2$C algorithm. We compared the Scaling-up Mix (part-1), the Multi-class Mix (part-2) and the Multi-jittering-class Mix (part-3) individually as augmentation methods with the baseline (MPL) by combining them and performing a comparative analysis. From Table~\ref{tab:table3}, it can be seen that each component contributes positively to the performance improvement of the segmentation model in different aspects. Part 1 focuses on improving the segmentation model's ability to detect the myocardium and left ventricle regions, while it doesn't show a significant effect on the contour detection ability of the ventricles and myocardium. Furthermore, the fifth row shows that the performance of the three segmentation objects decreases without Part 1.  In the second row, Multi-class Mix (part-2) provides the segmentation model with better region and contour recognition performance
. In the fourth row, the results indicate that introducing deformation operations, which is part-3,  can have a positive impact on the model's learning of semantic features, especially contours. In the last row of the table, the results demonstrate the necessity of combining the above three components. It achieves superior segmentation performance for all catego-ries (Mean DSC: 90.18\%, Mean HD: 2.96mm), proving the rationale behind the combination of these three components to enhance the baseline segmentation accuracy and robustness.

To indicate the effectiveness of the Scaling-up, we compare the performances between it and increasing batch size. We train the proposed framework without the Scaling-up and set the batch size as 48 so that the amount of inputs in an iteration is the same in both situations. Table~\ref{tab:table8} shows that the Scaling-up has a better ability to utilize more samples for a parameter update.

\begin{table*}[b]
  \centering
  \setlength{\tabcolsep}{4.0pt}
  \fontsize{7.5}{10.5}\selectfont
  \caption{Ablation study of components in SM$^2$C on the ACDC}\label{tab:table3}
    \begin{tabular}{ccccccccccc}
    \toprule[1pt]
    \multicolumn{3}{c}{Method}&\multicolumn{2}{c}{RV}&\multicolumn{2}{c}{Myo}&\multicolumn{2}{c}{LV}&\multicolumn{2}{c}{Mean}\cr
    \cmidrule(lr){1-11}
    {$\rm part\text{-}1$}&{$\rm part\text{-}2$}&{$\rm part\text{-}3$}&{$\rm DSC(\%)$}&{$\rm HD_{95}(mm)$} &{$\rm DSC(\%)$}&{$\rm HD_{95}(mm)$} &{$\rm DSC(\%)$}&{$\rm HD_{95}(mm)$} &{$\rm DSC(\%)$}&{$\rm HD_{95}(mm)$}\cr
    \midrule
    &&&83.47(0.07)&2.19(1.71)&82.17(0.04)&2.48(5.41)&88.69(0.08)&5.01(12.28)&84.77(0.06)&3.23(6.47)\cr
    \checkmark&&&79.98(0.14)&4.83(8.54)&83.91(0.05)&8.7(18.07)&88.72(0.1)&13.07(22.55)&84.2(0.1)&8.86(16.39)\cr
    &\checkmark&&87.83(0.07)&1.75(1.61)&84.31(0.04)&1.84(2.08)&91.31(0.07)&4.66(9.55)&87.82(0.06)&2.75(4.42)\cr
    \checkmark&\checkmark&&89.01(0.06)&1.62(1.83)&86.11(0.04)&2.46(4.09)&90.99(0.09)&4.24(8.31)&88.70(0.06)&2.78(4.75)\cr
    &\checkmark&\checkmark&89.08(0.07)&1.5(1.0)&85.07(0.04)&7.31(15.88)&91.11(0.07)&7.48(13.2)&88.42(0.88)&5.43(5.43)\cr
    \checkmark&\checkmark&\checkmark&{\bf 90.19(0.06)}&{\bf 1.29(0.83)}&{\bf 87.33(0.04)}&{\bf 1.19(0.31)}&{\bf 92.93(0.06)}&{\bf 2.39(4.80)}&{\bf 90.15(0.05)}&{\bf 1.62(1.98)}\cr
    \bottomrule[1pt]
    \end{tabular}
\end{table*}

\begin{table}[width=.9\linewidth,cols=3,pos=h]
\fontsize{7.5}{10.5}\selectfont
\caption{Comparison between the operations of Increasing Batch-Size and Scaling-up.}\label{tab:table8}
\begin{tabular*}{\tblwidth}{@{} CCC@{} }
\toprule
{$\rm Method$} & {$\rm Mean\ DSC(\%)$} & {$\rm Mean\ HD_{95}(mm)$}\\
\midrule
Batch-Size & 83.75 & 9.91\\
Scaling-up & 90.15 & 1.62\\
\bottomrule
\end{tabular*}
\end{table}

We'll also discuss the number of mixed segmentation objects in the augmented images. Since the proposed method performs augmentation by mixing segmentation objects from four images, we introduce the hyperparameter $\sigma$ to represent the number of mixed segmentation object groups in each image involved in the Multi-class mix. In each image, besides its own segmentation object group (i.e. $\sigma=1$), we can also mix parts of segmentation object groups from the other three images. As shown in Table~\ref{tab:table4}, when $\sigma=4$, indicating that each scaled-up image in SM$^2$C contains four groups of segmentation objects, SM$^2$C exhibits the best regional segmentation performance and can identify the most object pixels for the RV and Myo classes. However, when $\sigma=1$, the segmentation results for ventricles and myocardium are weaker than the baseline. Regarding the HD values, the results indicate that when $\sigma=2$, the framework excels in boundary detection and achieves segmentation contours that are closest to the ground truth. This suggests that increasing the number of segmentation objects does not consistently lead to a positive optimization of the model. Therefore, the choice of an appropriate $\sigma$ value should be task-specific.
\begin{table*}[h]
  \centering
  \setlength{\tabcolsep}{4.0pt}
  \fontsize{7.5}{10.5}\selectfont
  \caption{Comparison of the number of objects for SM$^2$C on the ACDC}\label{tab:table4}
    \begin{tabular}{lcccccccc}
    \toprule[1pt]
    \multirow{2}{*}{Method}&
    \multicolumn{2}{c}{RV}&\multicolumn{2}{c}{Myo}&\multicolumn{2}{c}{LV}&\multicolumn{2}{c}{Mean}\cr
    \cmidrule(lr){2-9}
    &{$\rm DSC(\%)$}&{$\rm HD_{95}(mm)$} &{$\rm DSC(\%)$}&{$\rm HD_{95}(mm)$} &{$\rm DSC(\%)$}&{$\rm HD_{95}(mm)$} &{$\rm DSC(\%)$}&{$\rm HD_{95}(mm)$}\cr
    \midrule
    
    Baseline&83.47(0.07)&2.19(1.71)&82.17(0.04)&2.48(5.41)&88.69(0.08)&5.01(12.28)&84.77(0.06)&3.23(6.47)\cr
    $\sigma=1$&79.98(0.14)&4.83(8.54)&83.91(0.05)&8.7(18.07)&88.72(0.1)&13.07(22.55)&84.2(0.1)&8.86(16.39)\cr
    $\sigma=2$&89.3(0.07)&1.4(0.81)&87.14(0.03)&1.65(3.0)&93.01(0.05)&3.8(7.98)&89.82(0.05)&2.28(3.93)\cr
    $\sigma=3$&88.01(0.09)&1.86(2.78)&86.16(0.04)&2.39(5.26)&91.3(0.07)&5.78(12.35)&88.49(0.07)&3.34(6.79)\cr
    $\sigma=4$&{\bf 90.19(0.06)}&{\bf 1.29(0.83)}&{\bf 87.33(0.04)}&{\bf 1.19(0.31)}&{\bf 92.93(0.06)}&{\bf 2.39(4.80)}&{\bf 90.15(0.05)}&{\bf 1.62(1.98)}\cr
    \bottomrule[1pt]
    \end{tabular}
\end{table*}

In addition to translation, stretching, and flipping, we also attempted to incorporate more augmentation methods to increase image diversity. On top of the SM$^2$C foundation, we introduced three additional data augmentation techniques to evaluate their performance: image deformation before input (technic-1), cutout method (technic-2), and color variation with Gaussian noise (technic-3). As shown in Table~\ref{tab:table5}, we observed that after incorporating these three techniques, the performance of the proposed algorithm decreased. It showed no advantage in either region or contour recognition. Furthermore, a similar phenomenon was observed in the SCGM and prostate datasets. This result suggests that combining SM$^2$C with these three techniques to improve data augmentation performance is not reasonable.

\begin{table*}[h]
  \centering
  \setlength{\tabcolsep}{4.0pt}
  \fontsize{7.5}{10.5}\selectfont
  \caption{Comparison between SM$^2$C combined with extra standard augmenting technics on the ACDC}\label{tab:table5}
    \begin{tabular}{lcccccccc}
    \toprule[1pt]
    \multirow{2}{*}{Method}&
    \multicolumn{2}{c}{RV}&\multicolumn{2}{c}{Myo}&\multicolumn{2}{c}{LV}&\multicolumn{2}{c}{Mean}\cr
    \cmidrule(lr){2-9}
    &{$\rm DSC(\%)$}&{$\rm HD_{95}(mm)$} &{$\rm DSC(\%)$}&{$\rm HD_{95}(mm)$} &{$\rm DSC(\%)$}&{$\rm HD_{95}(mm)$} &{$\rm DSC(\%)$}&{$\rm HD_{95}(mm)$}\cr
    \midrule
    w\textbackslash o technics&90.19(0.06)&1.29(0.83)&87.33(0.04)&1.19(0.31)&92.93(0.06)&2.39(4.80)&90.15(0.05)&1.62(1.98)\cr
    technic-1&84.87(0.11)&2.24(2.29)&84.73(0.04)&4.33(9.18)&89.97(0.08)&9.27(15.77)&86.52(0.08)&5.28(9.08)\cr
    technic-2&87.79(0.06)&1.88(1.93)&85.12(0.05)&3.08(7.68)&90.47(0.08)&6.16(13.66)&87.8(0.06)&3.71(7.76)\cr
    technic-3&86.68(0.09)&1.57(1.02)&84.94(0.05)&3.09(7.63)&91.06(0.07)&5.75(13.43)&87.56(0.07)&3.47(7.36)\cr
    \bottomrule[1pt]
    \end{tabular}
\end{table*}

Table~\ref{tab:table6} shows the segmentation performance of neural networks on the left and right ventricles when consistency regularisation in MPL is combined with different perturbation methods. Overall, the normal data augmentation methods, cutmix, cowmix and classmix all show varying degrees of improvement over the baseline (MPL). Among them, classmix performs best. Building on this, the segmentation model trained with the SM$^2$C algorithm shows a better DSC average, with improvements of 7.68\% and 0.84\% in 3 and 7 cases, respectively, compared to ClassMix. The improvements are impressive, especially in 3 cases, but the advantage of the proposed algorithm in terms of HD average is not as pronounced. In 3 cases, there is a slight decrease in the ability to recognise contours, especially in the recognition of the contours of the left ventricle and the myocardium. We speculate that the morphological variations of the left ventricle and myocardium from the base to the apex are significant, and that the strong augmentation capability of our method inadvertently affects the learning of semantic information about these contours by the neural network.

\begin{table*}[h]
  \centering
  \setlength{\tabcolsep}{4pt}
  \fontsize{7}{10.5}\selectfont
  \caption{Comparison between the proposed method and other data augmentation methods on the ACDC dataset.}\label{tab:table6}
    \begin{tabular}{clcccccccc}
    \toprule[1pt]
    \multicolumn{2}{c}{\multirow{2}{*}{Methods}}&
    \multicolumn{2}{c}{RV}&\multicolumn{2}{c}{Myo}&\multicolumn{2}{c}{LV}&\multicolumn{2}{c}{Mean}\cr
    \cmidrule(lr){3-10}
    &&{$\rm DSC(\%)$}&{$\rm HD_{95}(mm)$} &{$\rm DSC(\%)$}&{$\rm HD_{95}(mm)$} &{$\rm DSC(\%)$}&{$\rm HD_{95}(mm)$} &{$\rm DSC(\%)$}&{$\rm HD_{95}(mm)$}\cr
    \midrule
    \multirow{6}{*}{3 cases}
    &Baseline&48.18(0.32)&21.06(22.76)&62.24(0.28)&11.51(17.4)&69.19(0.3)&15.13(21.89)&59.87(0.3)&15.90(20.69)\cr
    &Normal&57.46(0.31)&14.54(18.57)&62.15(0.28)&13.89(18.54)&70.97(0.3)&19.89(23.29)&63.53(0.3)&16.11(20.14)\cr
    &CutMix&54.56(0.32)&14.55(18.42)&60.54(0.29)&11.76(17.87)&68.33(0.32)&14.49(22.64)&61.14(0.31)&13.6(19.64)\cr
    &CowMix&61.4(0.32)&12.97(19.52)&67.27(0.27)&5.71(9.82)&78.67(0.25)&4.0(5.68)&69.11(0.28)&7.56(11.67)\cr
    &ClassMix&72.21(0.28)&7.3(13.45)&72.44(0.22)&3.44(5.74)&84.78(0.17)&{\bf 5.15(11.55)}&76.48(0.23)&5.3(10.25)\cr
    &Proposed&{\bf 84.69(0.08)}&{\bf 4.22(7.57)}&{\bf 82.96(0.05)}&{\bf 3.41(5.29)}&{\bf 87.93(0.10)}&6.79(10.20)&{\bf 85.19(0.08)}&{\bf 4.80(7.69)}\cr
    \cmidrule(lr){2-10}
    \cmidrule(lr){1-10}
    \multirow{6}{*}{7 cases}
    &Baseline&83.47(0.1)&3.86(8.44)&82.17(0.05)&6.10(11.94)&88.69(0.09)&7.95(15.82)&84.77(0.08)&5.97(12.07)\cr
    &Normal&84.49(0.07)&2.19(1.71)&85.64(0.04)&2.48(5.41)&90.96(0.08)&5.01(12.28)&87.03(0.06)&3.23(6.47)\cr
    &CutMix&84.78(0.07)&2.15(1.68)&84.9(0.04)&3.73(6.54)&89.12(0.09)&9.5(15.31)&86.27(0.07)&5.13(7.84)\cr
    &CowMix&85.28(0.08)&5.57(15.71)&84.26(0.04)&3.68(9.01)&89.81(0.09)&6.8(14.05)&86.45(0.07)&5.35(12.93)\cr
    &ClassMix&89.11(0.07)&1.52(1.20)&86.05(0.04)&4.52(14.23)&92.85(0.06)&3.18(8.69)&89.34(0.06)&3.08(8.04)\cr
    &Proposed&{\bf 90.19(0.06)}&{\bf 1.29(0.83)}&{\bf 87.33(0.04)}&{\bf 1.19(0.31)}&{\bf 92.93(0.06)}&{\bf 2.39(4.80)}&{\bf 90.15(0.05)}&{\bf 1.62(1.98)}\cr
    \cmidrule(lr){1-10}
    &FS&90.81(0.06)&1.67(3.33)&89.46(0.03)&1.36(2.19)&94.93(0.03)&1.87(5.3)&91.73(0.04)&1.63(3.61) \cr
    \bottomrule[1pt]
    \end{tabular}
\end{table*}

\section{Conclusion}
In this paper, we have proposed a novel data augmentation method called SM$^2$C and combined it with a popular semi-supervised framework for semantic segmentation of medical images. SM$^2$C generates augmented images and artificial labels by mixing multiple unlabelled images using the network's semantic predictions. The generated images have different object shapes and richer foreground-background combinations, allowing the network to pay more attention to object boundaries. Our evaluation on three datasets demonstrated its superiority over existing methods and showed that it improves the state of the art. Its potential application in scenarios with limited medical images is noteworthy. Finally, ablation studies proved the rationality of SM$^2$C by comparing different configurations.


\bibliographystyle{cas-model2-names}

\bibliography{cas-refs}

\begin{thebibliography}{50}
\expandafter\ifx\csname natexlab\endcsname\relax\def\natexlab#1{#1}\fi
\providecommand{\url}[1]{\texttt{#1}}
\providecommand{\href}[2]{#2}
\providecommand{\path}[1]{#1}
\providecommand{\DOIprefix}{doi:}
\providecommand{\ArXivprefix}{arXiv:}
\providecommand{\URLprefix}{URL: }
\providecommand{\Pubmedprefix}{pmid:}
\providecommand{\doi}[1]{\href{http://dx.doi.org/#1}{\path{#1}}}
\providecommand{\Pubmed}[1]{\href{pmid:#1}{\path{#1}}}
\providecommand{\bibinfo}[2]{#2}
\ifx\xfnm\relax \def\xfnm[#1]{\unskip,\space#1}\fi
\bibitem[{Badrinarayanan et~al.(2017)Badrinarayanan, Kendall and Cipolla}]{r3}
\bibinfo{author}{Badrinarayanan, V.}, \bibinfo{author}{Kendall, A.}, \bibinfo{author}{Cipolla, R.}, \bibinfo{year}{2017}.
\newblock \bibinfo{title}{Segnet: A deep convolutional encoder-decoder architecture for image segmentation}.
\newblock \bibinfo{journal}{IEEE transactions on pattern analysis and machine intelligence} \bibinfo{volume}{39}, \bibinfo{pages}{2481--2495}.
\bibitem[{Bai et~al.(2017)Bai, Oktay, Sinclair, Suzuki, Rajchl, Tarroni, Glocker, King, Matthews and Rueckert}]{r34}
\bibinfo{author}{Bai, W.}, \bibinfo{author}{Oktay, O.}, \bibinfo{author}{Sinclair, M.}, \bibinfo{author}{Suzuki, H.}, \bibinfo{author}{Rajchl, M.}, \bibinfo{author}{Tarroni, G.}, \bibinfo{author}{Glocker, B.}, \bibinfo{author}{King, A.}, \bibinfo{author}{Matthews, P.M.}, \bibinfo{author}{Rueckert, D.}, \bibinfo{year}{2017}.
\newblock \bibinfo{title}{Semi-supervised learning for network-based cardiac mr image segmentation}, in: \bibinfo{booktitle}{MICCAI}, \bibinfo{organization}{Springer}. pp. \bibinfo{pages}{253--260}.
\bibitem[{Basak et~al.(2022)Basak, Bhattacharya, Hussain and Chatterjee}]{r20}
\bibinfo{author}{Basak, H.}, \bibinfo{author}{Bhattacharya, R.}, \bibinfo{author}{Hussain, R.}, \bibinfo{author}{Chatterjee, A.}, \bibinfo{year}{2022}.
\newblock \bibinfo{title}{An embarrassingly simple consistency regularization method for semi-supervised medical image segmentation}.
\newblock \bibinfo{journal}{arXiv preprint arXiv:2202.00677} .
\bibitem[{Baumgartner et~al.(2018)Baumgartner, Koch, Pollefeys and Konukoglu}]{r4}
\bibinfo{author}{Baumgartner, C.F.}, \bibinfo{author}{Koch, L.M.}, \bibinfo{author}{Pollefeys, M.}, \bibinfo{author}{Konukoglu, E.}, \bibinfo{year}{2018}.
\newblock \bibinfo{title}{An exploration of 2d and 3d deep learning techniques for cardiac mr image segmentation}, in: \bibinfo{booktitle}{STACOM}, \bibinfo{organization}{Springer}. pp. \bibinfo{pages}{111--119}.
\bibitem[{Bernard et~al.(2018)Bernard, Lalande, Zotti, Cervenansky, Yang, Heng, Cetin, Lekadir, Camara, Ballester et~al.}]{r32}
\bibinfo{author}{Bernard, O.}, \bibinfo{author}{Lalande, A.}, \bibinfo{author}{Zotti, C.}, \bibinfo{author}{Cervenansky, F.}, \bibinfo{author}{Yang, X.}, \bibinfo{author}{Heng, P.A.}, \bibinfo{author}{Cetin, I.}, \bibinfo{author}{Lekadir, K.}, \bibinfo{author}{Camara, O.}, \bibinfo{author}{Ballester, M.A.G.}, et~al., \bibinfo{year}{2018}.
\newblock \bibinfo{title}{Deep learning techniques for automatic mri cardiac multi-structures segmentation and diagnosis: is the problem solved?}
\newblock \bibinfo{journal}{IEEE transactions on medical imaging} \bibinfo{volume}{37}, \bibinfo{pages}{2514--2525}.
\bibitem[{Berthelot et~al.(2019)Berthelot, Carlini, Goodfellow, Papernot, Oliver and Raffel}]{r19}
\bibinfo{author}{Berthelot, D.}, \bibinfo{author}{Carlini, N.}, \bibinfo{author}{Goodfellow, I.}, \bibinfo{author}{Papernot, N.}, \bibinfo{author}{Oliver, A.}, \bibinfo{author}{Raffel, C.A.}, \bibinfo{year}{2019}.
\newblock \bibinfo{title}{Mixmatch: A holistic approach to semi-supervised learning}.
\newblock \bibinfo{journal}{Advances in neural information processing systems} \bibinfo{volume}{32}.
\bibitem[{Bortsova et~al.(2019)Bortsova, Dubost, Hogeweg, Katramados and De~Bruijne}]{r17}
\bibinfo{author}{Bortsova, G.}, \bibinfo{author}{Dubost, F.}, \bibinfo{author}{Hogeweg, L.}, \bibinfo{author}{Katramados, I.}, \bibinfo{author}{De~Bruijne, M.}, \bibinfo{year}{2019}.
\newblock \bibinfo{title}{Semi-supervised medical image segmentation via learning consistency under transformations}, in: \bibinfo{booktitle}{MICCAI}, \bibinfo{organization}{Springer}. pp. \bibinfo{pages}{810--818}.
\bibitem[{Chen et~al.(2022a)Chen, Fu, Xie, Zheng, Geng and Sham}]{r14}
\bibinfo{author}{Chen, J.}, \bibinfo{author}{Fu, C.}, \bibinfo{author}{Xie, H.}, \bibinfo{author}{Zheng, X.}, \bibinfo{author}{Geng, R.}, \bibinfo{author}{Sham, C.W.}, \bibinfo{year}{2022}a.
\newblock \bibinfo{title}{Uncertainty teacher with dense focal loss for semi-supervised medical image segmentation}.
\newblock \bibinfo{journal}{Computers in Biology and Medicine} \bibinfo{volume}{149}, \bibinfo{pages}{106034}.
\bibitem[{Chen et~al.(2022b)Chen, Fu, Xie, Zheng, Geng and Sham}]{r46}
\bibinfo{author}{Chen, J.}, \bibinfo{author}{Fu, C.}, \bibinfo{author}{Xie, H.}, \bibinfo{author}{Zheng, X.}, \bibinfo{author}{Geng, R.}, \bibinfo{author}{Sham, C.W.}, \bibinfo{year}{2022}b.
\newblock \bibinfo{title}{Uncertainty teacher with dense focal loss for semi-supervised medical image segmentation}.
\newblock \bibinfo{journal}{Computers in Biology and Medicine} \bibinfo{volume}{149}, \bibinfo{pages}{106034}.
\bibitem[{Chen et~al.(2021)Chen, Yuan, Zeng and Wang}]{r11}
\bibinfo{author}{Chen, X.}, \bibinfo{author}{Yuan, Y.}, \bibinfo{author}{Zeng, G.}, \bibinfo{author}{Wang, J.}, \bibinfo{year}{2021}.
\newblock \bibinfo{title}{Semi-supervised semantic segmentation with cross pseudo supervision}, in: \bibinfo{booktitle}{Proceedings of the IEEE/CVF Conference on Computer Vision and Pattern Recognition}, pp. \bibinfo{pages}{2613--2622}.
\bibitem[{Cubuk et~al.(2019)Cubuk, Zoph, Mane, Vasudevan and Le}]{r23}
\bibinfo{author}{Cubuk, E.D.}, \bibinfo{author}{Zoph, B.}, \bibinfo{author}{Mane, D.}, \bibinfo{author}{Vasudevan, V.}, \bibinfo{author}{Le, Q.V.}, \bibinfo{year}{2019}.
\newblock \bibinfo{title}{Autoaugment: Learning augmentation strategies from data}, in: \bibinfo{booktitle}{Proceedings of the IEEE/CVF conference on computer vision and pattern recognition}, pp. \bibinfo{pages}{113--123}.
\bibitem[{Cubuk et~al.(2020)Cubuk, Zoph, Shlens and Le}]{r24}
\bibinfo{author}{Cubuk, E.D.}, \bibinfo{author}{Zoph, B.}, \bibinfo{author}{Shlens, J.}, \bibinfo{author}{Le, Q.V.}, \bibinfo{year}{2020}.
\newblock \bibinfo{title}{Randaugment: Practical automated data augmentation with a reduced search space}, in: \bibinfo{booktitle}{Proceedings of the IEEE/CVF conference on computer vision and pattern recognition workshops}, pp. \bibinfo{pages}{702--703}.
\bibitem[{French et~al.(2019)French, Laine, Aila, Mackiewicz and Finlayson}]{r16}
\bibinfo{author}{French, G.}, \bibinfo{author}{Laine, S.}, \bibinfo{author}{Aila, T.}, \bibinfo{author}{Mackiewicz, M.}, \bibinfo{author}{Finlayson, G.}, \bibinfo{year}{2019}.
\newblock \bibinfo{title}{Semi-supervised semantic segmentation needs strong, varied perturbations}.
\newblock \bibinfo{journal}{arXiv preprint arXiv:1906.01916} .
\bibitem[{French et~al.(2020)French, Oliver and Salimans}]{r27}
\bibinfo{author}{French, G.}, \bibinfo{author}{Oliver, A.}, \bibinfo{author}{Salimans, T.}, \bibinfo{year}{2020}.
\newblock \bibinfo{title}{Milking cowmask for semi-supervised image classification}.
\newblock \bibinfo{journal}{arXiv preprint arXiv:2003.12022} .
\bibitem[{Han et~al.(2022)Han, Liu, Song, Liu, Qiu, Tang, Teng and Liu}]{r13}
\bibinfo{author}{Han, K.}, \bibinfo{author}{Liu, L.}, \bibinfo{author}{Song, Y.}, \bibinfo{author}{Liu, Y.}, \bibinfo{author}{Qiu, C.}, \bibinfo{author}{Tang, Y.}, \bibinfo{author}{Teng, Q.}, \bibinfo{author}{Liu, Z.}, \bibinfo{year}{2022}.
\newblock \bibinfo{title}{An effective semi-supervised approach for liver ct image segmentation}.
\newblock \bibinfo{journal}{IEEE Journal of Biomedical and Health Informatics} \bibinfo{volume}{26}, \bibinfo{pages}{3999--4007}.
\bibitem[{He et~al.(2016)He, Zhang, Ren and Sun}]{r51}
\bibinfo{author}{He, K.}, \bibinfo{author}{Zhang, X.}, \bibinfo{author}{Ren, S.}, \bibinfo{author}{Sun, J.}, \bibinfo{year}{2016}.
\newblock \bibinfo{title}{Deep residual learning for image recognition}, in: \bibinfo{booktitle}{Proceedings of the IEEE conference on computer vision and pattern recognition}, pp. \bibinfo{pages}{770--778}.
\bibitem[{Hinton et~al.(2012)Hinton, Srivastava and Swersky}]{r40}
\bibinfo{author}{Hinton, G.}, \bibinfo{author}{Srivastava, N.}, \bibinfo{author}{Swersky, K.}, \bibinfo{year}{2012}.
\newblock \bibinfo{title}{Neural networks for machine learning lecture 6a overview of mini-batch gradient descent}.
\newblock \bibinfo{journal}{Cited on} \bibinfo{volume}{14}, \bibinfo{pages}{2}.
\bibitem[{Hu et~al.(2022)Hu, Li, Peng, Xiao, Zhan, Zu, Wu, Zhou and Wang}]{r31}
\bibinfo{author}{Hu, L.}, \bibinfo{author}{Li, J.}, \bibinfo{author}{Peng, X.}, \bibinfo{author}{Xiao, J.}, \bibinfo{author}{Zhan, B.}, \bibinfo{author}{Zu, C.}, \bibinfo{author}{Wu, X.}, \bibinfo{author}{Zhou, J.}, \bibinfo{author}{Wang, Y.}, \bibinfo{year}{2022}.
\newblock \bibinfo{title}{Semi-supervised npc segmentation with uncertainty and attention guided consistency}.
\newblock \bibinfo{journal}{Knowledge-Based Systems} \bibinfo{volume}{239}, \bibinfo{pages}{108021}.
\bibitem[{Huang et~al.(2022)Huang, Chen, Xiong, Zhang, Chen, Sun and Wu}]{r18}
\bibinfo{author}{Huang, W.}, \bibinfo{author}{Chen, C.}, \bibinfo{author}{Xiong, Z.}, \bibinfo{author}{Zhang, Y.}, \bibinfo{author}{Chen, X.}, \bibinfo{author}{Sun, X.}, \bibinfo{author}{Wu, F.}, \bibinfo{year}{2022}.
\newblock \bibinfo{title}{Semi-supervised neuron segmentation via reinforced consistency learning}.
\newblock \bibinfo{journal}{IEEE Transactions on Medical Imaging} \bibinfo{volume}{41}, \bibinfo{pages}{3016--3028}.
\bibitem[{Jiao et~al.(2022)Jiao, Zhang, Ding, Cai and Zhang}]{r7}
\bibinfo{author}{Jiao, R.}, \bibinfo{author}{Zhang, Y.}, \bibinfo{author}{Ding, L.}, \bibinfo{author}{Cai, R.}, \bibinfo{author}{Zhang, J.}, \bibinfo{year}{2022}.
\newblock \bibinfo{title}{Learning with limited annotations: a survey on deep semi-supervised learning for medical image segmentation}.
\newblock \bibinfo{journal}{arXiv preprint arXiv:2207.14191} .
\bibitem[{Keskar et~al.(2016)Keskar, Mudigere, Nocedal, Smelyanskiy and Tang}]{r39}
\bibinfo{author}{Keskar, N.S.}, \bibinfo{author}{Mudigere, D.}, \bibinfo{author}{Nocedal, J.}, \bibinfo{author}{Smelyanskiy, M.}, \bibinfo{author}{Tang, P.T.P.}, \bibinfo{year}{2016}.
\newblock \bibinfo{title}{On large-batch training for deep learning: Generalization gap and sharp minima}.
\newblock \bibinfo{journal}{arXiv preprint arXiv:1609.04836} .
\bibitem[{Lee et~al.(2013)}]{r8}
\bibinfo{author}{Lee, D.H.}, et~al., \bibinfo{year}{2013}.
\newblock \bibinfo{title}{Pseudo-label: The simple and efficient semi-supervised learning method for deep neural networks}, in: \bibinfo{booktitle}{Workshop on challenges in representation learning, ICML}, p. \bibinfo{pages}{896}.
\bibitem[{Liu et~al.()Liu, Lin, Cao, Hu, Wei, Zhang, Lin and Guo}]{r45}
\bibinfo{author}{Liu, Z.}, \bibinfo{author}{Lin, Y.}, \bibinfo{author}{Cao, Y.}, \bibinfo{author}{Hu, H.}, \bibinfo{author}{Wei, Y.}, \bibinfo{author}{Zhang, Z.}, \bibinfo{author}{Lin, S.}, \bibinfo{author}{Guo, B.}, .
\newblock \bibinfo{title}{Swin transformer: Hierarchical vision transformer using shifted windows}, in: \bibinfo{booktitle}{ICCV}, \bibinfo{publisher}{{IEEE}}. pp. \bibinfo{pages}{9992--10002}.
\bibitem[{Long et~al.(2015)Long, Shelhamer and Darrell}]{r2}
\bibinfo{author}{Long, J.}, \bibinfo{author}{Shelhamer, E.}, \bibinfo{author}{Darrell, T.}, \bibinfo{year}{2015}.
\newblock \bibinfo{title}{Fully convolutional networks for semantic segmentation}, in: \bibinfo{booktitle}{Proceedings of the IEEE conference on computer vision and pattern recognition}, pp. \bibinfo{pages}{3431--3440}.
\bibitem[{Lu et~al.(2024)Lu, Yan, Chen, Cheng, Zhang and Yang}]{r47}
\bibinfo{author}{Lu, S.}, \bibinfo{author}{Yan, Z.}, \bibinfo{author}{Chen, W.}, \bibinfo{author}{Cheng, T.}, \bibinfo{author}{Zhang, Z.}, \bibinfo{author}{Yang, G.}, \bibinfo{year}{2024}.
\newblock \bibinfo{title}{Dual consistency regularization with subjective logic for semi-supervised medical image segmentation}.
\newblock \bibinfo{journal}{Computers in Biology and Medicine} \bibinfo{volume}{170}, \bibinfo{pages}{107991}.
\bibitem[{Luo et~al.(2021a)Luo, Chen, Song and Wang}]{r43}
\bibinfo{author}{Luo, X.}, \bibinfo{author}{Chen, J.}, \bibinfo{author}{Song, T.}, \bibinfo{author}{Wang, G.}, \bibinfo{year}{2021}a.
\newblock \bibinfo{title}{Semi-supervised medical image segmentation through dual-task consistency}, pp. \bibinfo{pages}{8801--8809}.
\bibitem[{Luo et~al.(2022a)Luo, Hu, Song, Wang and Zhang}]{r22}
\bibinfo{author}{Luo, X.}, \bibinfo{author}{Hu, M.}, \bibinfo{author}{Song, T.}, \bibinfo{author}{Wang, G.}, \bibinfo{author}{Zhang, S.}, \bibinfo{year}{2022}a.
\newblock \bibinfo{title}{Semi-supervised medical image segmentation via cross teaching between cnn and transformer}, in: \bibinfo{booktitle}{International Conference on Medical Imaging with Deep Learning}, \bibinfo{organization}{PMLR}. pp. \bibinfo{pages}{820--833}.
\bibitem[{Luo et~al.(2021b)Luo, Liao, Chen, Song, Chen, Zhang, Chen, Wang and Zhang}]{r42}
\bibinfo{author}{Luo, X.}, \bibinfo{author}{Liao, W.}, \bibinfo{author}{Chen, J.}, \bibinfo{author}{Song, T.}, \bibinfo{author}{Chen, Y.}, \bibinfo{author}{Zhang, S.}, \bibinfo{author}{Chen, N.}, \bibinfo{author}{Wang, G.}, \bibinfo{author}{Zhang, S.}, \bibinfo{year}{2021}b.
\newblock \bibinfo{title}{Efficient semi-supervised gross target volume of nasopharyngeal carcinoma segmentation via uncertainty rectified pyramid consistency}, in: \bibinfo{booktitle}{Medical Image Computing and Computer Assisted Intervention -- MICCAI 2021}, pp. \bibinfo{pages}{318--329}.
\bibitem[{Luo et~al.(2022b)Luo, Wang, Liao, Chen, Song, Chen, Zhang, Metaxas and Zhang}]{r41}
\bibinfo{author}{Luo, X.}, \bibinfo{author}{Wang, G.}, \bibinfo{author}{Liao, W.}, \bibinfo{author}{Chen, J.}, \bibinfo{author}{Song, T.}, \bibinfo{author}{Chen, Y.}, \bibinfo{author}{Zhang, S.}, \bibinfo{author}{Metaxas, D.N.}, \bibinfo{author}{Zhang, S.}, \bibinfo{year}{2022}b.
\newblock \bibinfo{title}{Semi-supervised medical image segmentation via uncertainty rectified pyramid consistency}.
\newblock \bibinfo{journal}{Medical Image Analysis} \bibinfo{volume}{80}, \bibinfo{pages}{102517}.
\bibitem[{Olsson et~al.()Olsson, Tranheden, Pinto and Svensson}]{r30}
\bibinfo{author}{Olsson, V.}, \bibinfo{author}{Tranheden, W.}, \bibinfo{author}{Pinto, J.}, \bibinfo{author}{Svensson, L.}, .
\newblock \bibinfo{title}{{ClassMix}: Segmentation-based data augmentation for semi-supervised learning}, in: \bibinfo{booktitle}{WACV}, \bibinfo{publisher}{IEEE}. pp. \bibinfo{pages}{1368--1377}.
\bibitem[{Ouali et~al.(2020)Ouali, Hudelot and Tami}]{r36}
\bibinfo{author}{Ouali, Y.}, \bibinfo{author}{Hudelot, C.}, \bibinfo{author}{Tami, M.}, \bibinfo{year}{2020}.
\newblock \bibinfo{title}{Semi-supervised semantic segmentation with cross-consistency training}, in: \bibinfo{booktitle}{Proceedings of the IEEE/CVF Conference on Computer Vision and Pattern Recognition}, pp. \bibinfo{pages}{12674--12684}.
\bibitem[{Peng et~al.(2020)Peng, Estrada, Pedersoli and Desrosiers}]{r6}
\bibinfo{author}{Peng, J.}, \bibinfo{author}{Estrada, G.}, \bibinfo{author}{Pedersoli, M.}, \bibinfo{author}{Desrosiers, C.}, \bibinfo{year}{2020}.
\newblock \bibinfo{title}{Deep co-training for semi-supervised image segmentation}.
\newblock \bibinfo{journal}{Pattern Recognition} \bibinfo{volume}{107}, \bibinfo{pages}{107269}.
\bibitem[{Pham et~al.(2021)Pham, Dai, Xie and Le}]{r15}
\bibinfo{author}{Pham, H.}, \bibinfo{author}{Dai, Z.}, \bibinfo{author}{Xie, Q.}, \bibinfo{author}{Le, Q.V.}, \bibinfo{year}{2021}.
\newblock \bibinfo{title}{Meta pseudo labels}, in: \bibinfo{booktitle}{Proceedings of the IEEE/CVF conference on computer vision and pattern recognition}, pp. \bibinfo{pages}{11557--11568}.
\bibitem[{Prados et~al.(2017)Prados, Ashburner, Blaiotta, Brosch, Carballido-Gamio, Cardoso, Conrad, Datta, Dávid, Leener, Dupont, Freund, Wheeler-Kingshott, Grussu, Henry, Landman, Ljungberg, Lyttle, Ourselin, Papinutto, Saporito, Schlaeger, Smith, Summers, Tam, Yiannakas, Zhu and Cohen-Adad}]{r33}
\bibinfo{author}{Prados, F.}, \bibinfo{author}{Ashburner, J.}, \bibinfo{author}{Blaiotta, C.}, \bibinfo{author}{Brosch, T.}, \bibinfo{author}{Carballido-Gamio, J.}, \bibinfo{author}{Cardoso, M.J.}, \bibinfo{author}{Conrad, B.N.}, \bibinfo{author}{Datta, E.}, \bibinfo{author}{Dávid, G.}, \bibinfo{author}{Leener, B.D.}, \bibinfo{author}{Dupont, S.M.}, \bibinfo{author}{Freund, P.}, \bibinfo{author}{Wheeler-Kingshott, C.A.G.}, \bibinfo{author}{Grussu, F.}, \bibinfo{author}{Henry, R.}, \bibinfo{author}{Landman, B.A.}, \bibinfo{author}{Ljungberg, E.}, \bibinfo{author}{Lyttle, B.}, \bibinfo{author}{Ourselin, S.}, \bibinfo{author}{Papinutto, N.}, \bibinfo{author}{Saporito, S.}, \bibinfo{author}{Schlaeger, R.}, \bibinfo{author}{Smith, S.A.}, \bibinfo{author}{Summers, P.}, \bibinfo{author}{Tam, R.}, \bibinfo{author}{Yiannakas, M.C.}, \bibinfo{author}{Zhu, A.}, \bibinfo{author}{Cohen-Adad, J.}, \bibinfo{year}{2017}.
\newblock \bibinfo{title}{Spinal cord grey matter segmentation challenge}.
\newblock \bibinfo{journal}{NeuroImage} \bibinfo{volume}{152}, \bibinfo{pages}{312--329}.
\bibitem[{Raj et~al.(2022)Raj, Mathew, Kannath and Rajan}]{r29}
\bibinfo{author}{Raj, R.}, \bibinfo{author}{Mathew, J.}, \bibinfo{author}{Kannath, S.K.}, \bibinfo{author}{Rajan, J.}, \bibinfo{year}{2022}.
\newblock \bibinfo{title}{Crossover based technique for data augmentation}.
\newblock \bibinfo{journal}{Computer Methods and Programs in Biomedicine} \bibinfo{volume}{218}, \bibinfo{pages}{106716}.
\bibitem[{Ronneberger et~al.(2015)Ronneberger, Fischer and Brox}]{r1}
\bibinfo{author}{Ronneberger, O.}, \bibinfo{author}{Fischer, P.}, \bibinfo{author}{Brox, T.}, \bibinfo{year}{2015}.
\newblock \bibinfo{title}{U-net: Convolutional networks for biomedical image segmentation}, in: \bibinfo{booktitle}{MICCAI}, \bibinfo{organization}{Springer}. pp. \bibinfo{pages}{234--241}.
\bibitem[{Simpson et~al.(2019)Simpson, Antonelli, Bakas, Bilello, Farahani, Van~Ginneken, Kopp-Schneider, Landman, Litjens, Menze et~al.}]{r35}
\bibinfo{author}{Simpson, A.L.}, \bibinfo{author}{Antonelli, M.}, \bibinfo{author}{Bakas, S.}, \bibinfo{author}{Bilello, M.}, \bibinfo{author}{Farahani, K.}, \bibinfo{author}{Van~Ginneken, B.}, \bibinfo{author}{Kopp-Schneider, A.}, \bibinfo{author}{Landman, B.A.}, \bibinfo{author}{Litjens, G.}, \bibinfo{author}{Menze, B.}, et~al., \bibinfo{year}{2019}.
\newblock \bibinfo{title}{A large annotated medical image dataset for the development and evaluation of segmentation algorithms}.
\newblock \bibinfo{journal}{arXiv preprint arXiv:1902.09063} .
\bibitem[{Tarvainen and Valpola(2017)}]{r12}
\bibinfo{author}{Tarvainen, A.}, \bibinfo{author}{Valpola, H.}, \bibinfo{year}{2017}.
\newblock \bibinfo{title}{Mean teachers are better role models: Weight-averaged consistency targets improve semi-supervised deep learning results}.
\newblock \bibinfo{journal}{Advances in neural information processing systems} \bibinfo{volume}{30}.
\bibitem[{Verma et~al.(2022)Verma, Kawaguchi, Lamb, Kannala, Solin, Bengio and Lopez-Paz}]{r37}
\bibinfo{author}{Verma, V.}, \bibinfo{author}{Kawaguchi, K.}, \bibinfo{author}{Lamb, A.}, \bibinfo{author}{Kannala, J.}, \bibinfo{author}{Solin, A.}, \bibinfo{author}{Bengio, Y.}, \bibinfo{author}{Lopez-Paz, D.}, \bibinfo{year}{2022}.
\newblock \bibinfo{title}{Interpolation consistency training for semi-supervised learning}.
\newblock \bibinfo{journal}{Neural Networks} \bibinfo{volume}{145}, \bibinfo{pages}{90--106}.
\bibitem[{Wang et~al.(2023)Wang, Peng, Pedersoli, Zhou, Zhang and Desrosiers}]{r49}
\bibinfo{author}{Wang, P.}, \bibinfo{author}{Peng, J.}, \bibinfo{author}{Pedersoli, M.}, \bibinfo{author}{Zhou, Y.}, \bibinfo{author}{Zhang, C.}, \bibinfo{author}{Desrosiers, C.}, \bibinfo{year}{2023}.
\newblock \bibinfo{title}{Cat: Constrained adversarial training for anatomically-plausible semi-supervised segmentation}.
\newblock \bibinfo{journal}{IEEE Transactions on Medical Imaging} .
\bibitem[{Wang et~al.(2022a)Wang, Ji and Xiao}]{r28}
\bibinfo{author}{Wang, Y.}, \bibinfo{author}{Ji, Y.}, \bibinfo{author}{Xiao, H.}, \bibinfo{year}{2022}a.
\newblock \bibinfo{title}{A data augmentation method for fully automatic brain tumor segmentation}.
\newblock \bibinfo{journal}{Computers in Biology and Medicine} \bibinfo{volume}{149}, \bibinfo{pages}{106039}.
\bibitem[{Wang et~al.(2022b)Wang, Li, Zheng and Huang}]{r50}
\bibinfo{author}{Wang, Z.}, \bibinfo{author}{Li, T.}, \bibinfo{author}{Zheng, J.Q.}, \bibinfo{author}{Huang, B.}, \bibinfo{year}{2022}b.
\newblock \bibinfo{title}{When cnn meet with vit: Towards semi-supervised learning for multi-class medical image semantic segmentation}, in: \bibinfo{booktitle}{European Conference on Computer Vision}, \bibinfo{organization}{Springer}. pp. \bibinfo{pages}{424--441}.
\bibitem[{Xie et~al.(2023)Xie, Fu, Zheng, Zheng, Sham and Wang}]{r48}
\bibinfo{author}{Xie, H.}, \bibinfo{author}{Fu, C.}, \bibinfo{author}{Zheng, X.}, \bibinfo{author}{Zheng, Y.}, \bibinfo{author}{Sham, C.W.}, \bibinfo{author}{Wang, X.}, \bibinfo{year}{2023}.
\newblock \bibinfo{title}{Adversarial co-training for semantic segmentation over medical images}.
\newblock \bibinfo{journal}{Computers in biology and medicine} \bibinfo{volume}{157}, \bibinfo{pages}{106736}.
\bibitem[{Xie et~al.(2020)Xie, Dai, Hovy, Luong and Le}]{r9}
\bibinfo{author}{Xie, Q.}, \bibinfo{author}{Dai, Z.}, \bibinfo{author}{Hovy, E.}, \bibinfo{author}{Luong, T.}, \bibinfo{author}{Le, Q.}, \bibinfo{year}{2020}.
\newblock \bibinfo{title}{Unsupervised data augmentation for consistency training}.
\newblock \bibinfo{journal}{Advances in neural information processing systems} \bibinfo{volume}{33}, \bibinfo{pages}{6256--6268}.
\bibitem[{Yu et~al.(2019)Yu, Wang, Li, Fu and Heng}]{r21}
\bibinfo{author}{Yu, L.}, \bibinfo{author}{Wang, S.}, \bibinfo{author}{Li, X.}, \bibinfo{author}{Fu, C.W.}, \bibinfo{author}{Heng, P.A.}, \bibinfo{year}{2019}.
\newblock \bibinfo{title}{Uncertainty-aware self-ensembling model for semi-supervised 3d left atrium segmentation}, in: \bibinfo{booktitle}{MICCAI}, \bibinfo{organization}{Springer}. pp. \bibinfo{pages}{605--613}.
\bibitem[{Yun et~al.(2019)Yun, Han, Oh, Chun, Choe and Yoo}]{r26}
\bibinfo{author}{Yun, S.}, \bibinfo{author}{Han, D.}, \bibinfo{author}{Oh, S.J.}, \bibinfo{author}{Chun, S.}, \bibinfo{author}{Choe, J.}, \bibinfo{author}{Yoo, Y.}, \bibinfo{year}{2019}.
\newblock \bibinfo{title}{Cutmix: Regularization strategy to train strong classifiers with localizable features}, in: \bibinfo{booktitle}{Proceedings of the IEEE/CVF international conference on computer vision}, pp. \bibinfo{pages}{6023--6032}.
\bibitem[{Zeng et~al.(2021)Zeng, Huang, Zhong, Sun, Han, Lin, Ni and Wang}]{r38}
\bibinfo{author}{Zeng, X.}, \bibinfo{author}{Huang, R.}, \bibinfo{author}{Zhong, Y.}, \bibinfo{author}{Sun, D.}, \bibinfo{author}{Han, C.}, \bibinfo{author}{Lin, D.}, \bibinfo{author}{Ni, D.}, \bibinfo{author}{Wang, Y.}, \bibinfo{year}{2021}.
\newblock \bibinfo{title}{Reciprocal learning for semi-supervised segmentation}, in: \bibinfo{booktitle}{Medical Image Computing and Computer Assisted Intervention--MICCAI 2021: 24th International Conference, Strasbourg, France, September 27--October 1, 2021, Proceedings, Part II 24}, \bibinfo{organization}{Springer}. pp. \bibinfo{pages}{352--361}.
\bibitem[{Zhang et~al.(2017)Zhang, Cisse, Dauphin and Lopez-Paz}]{r25}
\bibinfo{author}{Zhang, H.}, \bibinfo{author}{Cisse, M.}, \bibinfo{author}{Dauphin, Y.N.}, \bibinfo{author}{Lopez-Paz, D.}, \bibinfo{year}{2017}.
\newblock \bibinfo{title}{mixup: Beyond empirical risk minimization}.
\newblock \bibinfo{journal}{arXiv preprint arXiv:1710.09412} .
\bibitem[{Zheng et~al.(2018)Zheng, Delingette, Duchateau and Ayache}]{r5}
\bibinfo{author}{Zheng, Q.}, \bibinfo{author}{Delingette, H.}, \bibinfo{author}{Duchateau, N.}, \bibinfo{author}{Ayache, N.}, \bibinfo{year}{2018}.
\newblock \bibinfo{title}{3\--{D} consistent and robust segmentation of cardiac images by deep learning with spatial propagation}.
\newblock \bibinfo{journal}{IEEE transactions on medical imaging} \bibinfo{volume}{37}, \bibinfo{pages}{2137--2148}.
\bibitem[{Zheng et~al.(2022)Zheng, Fu, Xie, Chen, Wang and Sham}]{r10}
\bibinfo{author}{Zheng, X.}, \bibinfo{author}{Fu, C.}, \bibinfo{author}{Xie, H.}, \bibinfo{author}{Chen, J.}, \bibinfo{author}{Wang, X.}, \bibinfo{author}{Sham, C.W.}, \bibinfo{year}{2022}.
\newblock \bibinfo{title}{Uncertainty-aware deep co-training for semi-supervised medical image segmentation}.
\newblock \bibinfo{journal}{Computers in Biology and Medicine} \bibinfo{volume}{149}, \bibinfo{pages}{106051}.

\end{thebibliography}


\end{document}